\documentclass[DTMColor]{ROB-New}

\usepackage{amssymb}
\usepackage{amsmath}
\usepackage{graphics}
\usepackage{lipsum}
\usepackage{mathtools}
\usepackage{cuted}
\usepackage{tabularx}
\usepackage{array}
\usepackage{footnote} 
\usepackage{multirow}
\usepackage{booktabs}
\usepackage{url}
\usepackage{stfloats}
\usepackage{lineno}
\usepackage{multirow}
\usepackage{pifont}
\usepackage{color}
\usepackage{framed}

\let\sss= \scriptscriptstyle
\usepackage{enumitem}
\usepackage[ruled,linesnumbered]{algorithm2e}

\SetKw{Continue}{continue}

\begin{document}



\jnlPage{1}{7}
\jyear{2025}
\jdoi{10.1017/xxxxx}

\title{A Novel Semi-Coupled Hierarchical Motion Planning Framework for Cooperative Transportation of Multiple Mobile Manipulators}

\author{Heng Zhang}
\author{Haoyi Song}
\author{Wenhang Liu}
\author{Xinjun Sheng}
\author{Zhenhua Xiong\hyperlink{corr}{*}}
\author{Xiangyang Zhu\hyperlink{corr}{*}}
\address{School of Mechanical Engineering, Shanghai Jiao Tong University, Shanghai, China}
\address{\hypertarget{corr}{*}Corresponding author. \email{mexiong@sjtu.edu.cn}}



\abstract{Multiple mobile manipulators show superiority in the tasks requiring mobility and dexterity compared with a single robot, especially when manipulating/transporting bulky objects. However, closed-chain of the system, redundancy of each mobile manipulator and obstacles in the environment bring challenges to the motion planning problem. In this paper, we propose a novel semi-coupled hierarchical framework (\emph{SCHF}), which decomposes the problem into two semi-coupled sub-problems. To be specific, the centralized layer plans the object’s motion first and then the decentralized layer independently explores the redundancy of each robot in real-time. A notable feature is that the lower bound of the redundancy constraint metric is ensured besides the closed-chain and obstacle-avoidance constraints in the centralized layer, which ensures the object's motion can be executed by each robot in the decentralized layer. Simulated results show that the success rate and time cost of \emph{SCHF} outperforms the fully centralized planner and fully decoupled hierarchical planner significantly. In addition, cluttered real-world experiments also show the feasibility of the \emph{SCHF} in the transportation tasks. A video clip in various scenarios can be found at \url{https://youtu.be/Y8ZrnspIuBg}.}

\maketitle

\section{Introduction}

\begin{figure}[ht]
  \centering
  \includegraphics[width=7.5cm]{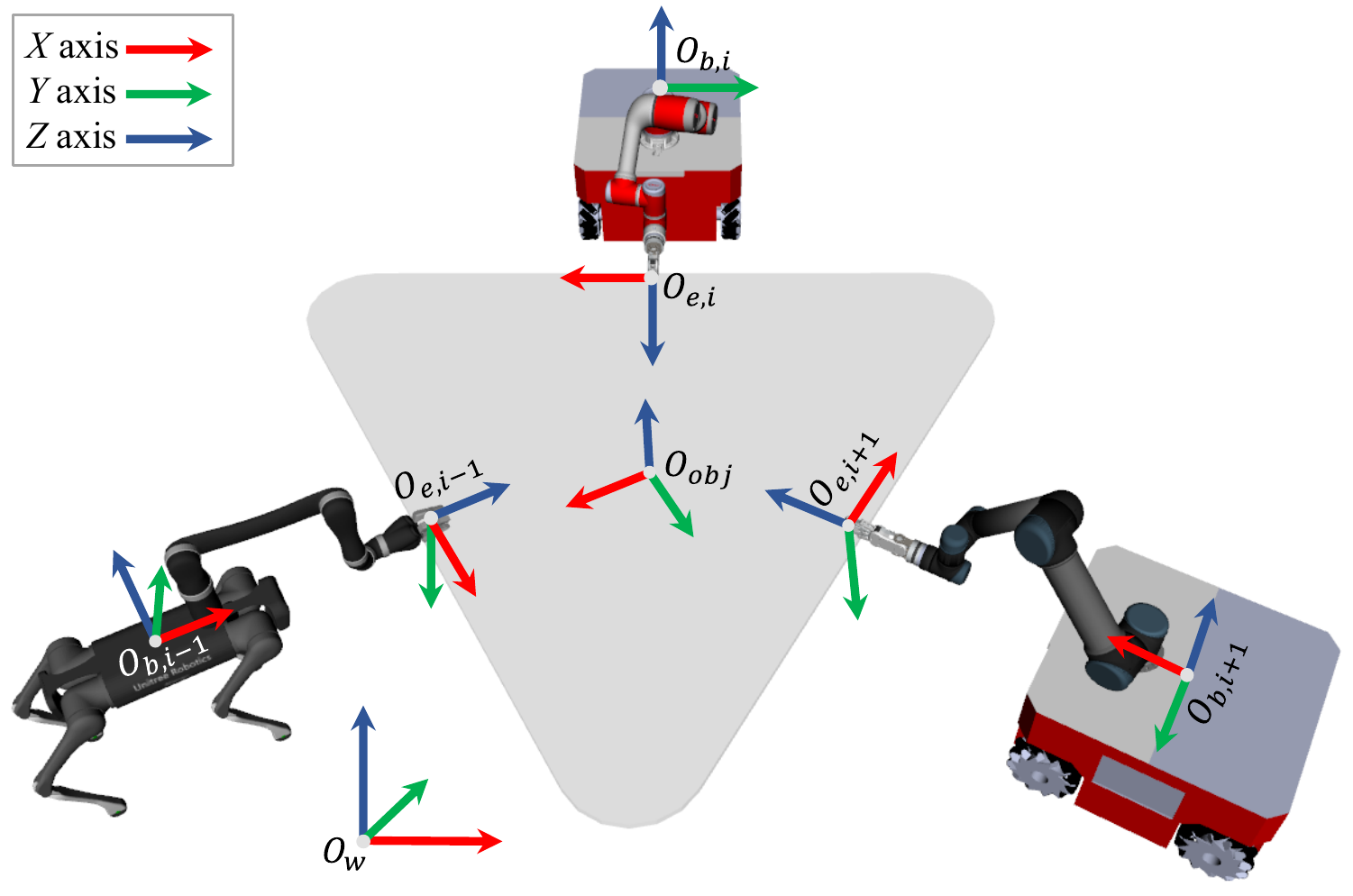}
  \caption{Multiple mobile manipulators system. The object should be transported while avoiding collision and obeying the closed-chain constraint.}
  \label{fig_robot_model}
\end{figure}

Multi-robot systems (\emph{MRS}) allows simpler, cheaper, modular robotic units to be reorganized into a group based on the task at hand. It can be as effective as a task-specific, larger, monolithic robot, which may be more expensive and has to be rebuilt according to the task. Inspired by nature, \emph{MRS} has evolved into a variety of forms, such as multiple mobile robots \cite{liu2023novel}, multiple manipulators \cite{suarez2018can}, and multiple drones \cite{chung2018survey}, and has been applied in various scenarios like construction \cite{petersen2019review} and transportation \cite{koung2021cooperative}.

Compared with a single robot, the fundamental challenge of the \emph{MRS} lies in the cooperation among robots. Cooperation may happen in several aspects, such as knowledge sharing and physical manipulation. In this paper, we focus on the motion planning of the \emph{MRS} when they fulfill some cooperative transportion tasks. 

As a combination of the mobile robot and the manipulator, mobile manipulator (\emph{MM}) inherits the mobility from the mobile robot and the dexterity from the manipulator. Compared with the two subsystems, \emph{MM} is redundant. This implies that the same task at the end effector can be executed in different ways in the configuration space, which gives the possibility of optimizing the joint configuration, such as avoiding obstacles. When the object and the robots are rigidly connected, closed-chain will form and the motion of the whole system will be restricted onto a lower-dimensional manifold, the connectivity of which will be further affected by various obstacles in the environment \cite{kingston2018sampling}. Therefore, closed-chain, redundancy and obstacle-avoidance are coupled and make the motion planning of multiple mobile manipulators complicated \cite{meng2024discrete, ebel2024cooperative, xu2023reinforcement}.

\subsection{Related Works} \label{section_related_work}

Redundancy of the \emph{MM} can be solved at joint position level \cite{ancona2017redundancy} or joint velocity level \cite{chiaverini1997singularity}. For the former, redundancy parameters will be designed to restrict the motion of the \emph{MM} \cite{ancona2017redundancy}. For the latter, redundancy can be dealt with by the task-priority algorithm \cite{chiaverini1997singularity}, where the additional task is only satisfied in the null space of the primary task and thus gives a higher priority to the primary task. In \cite{zhang2019task}, the motion of the end effector and the mobile robot were treated as the primary task and the additional task, respectively. The experimental results showed that the manipulator could grasp the object while the mobile robot moving.  The key to the redundancy resolution is fulfilling additional constraints while obeying the end effector's motion. For example, our previous work \cite{zhang2022cooperative} considered the dexterity of the \emph{MM}, obstacle-avoidance of the mobile robot and formation of the system and realized intuitional human-robot cooperative transportation task.

Closed-chain constrained manifold has its dimension lower than that of the configuration space, hence uniform sampling in the configuration space has zero probability of generating a valid state that satisfies this constraint \cite{kingston2018sampling}. One solution to this problem is introducing an allowable tolerance \cite{bonilla2017noninteracting}. Therefore, the volume of the satisfying subset grows, and various sampling-based motion planners for the unconstrained problem can be adopted. However, much of the complexity of handling the constraint transfers from the higher-level motion planner to the lower-level motion controller. In the case that multiple robots are rigidly connected, motion deviation caused by constraint relaxation may damage the robots and the transported object.

Another popular approach to resolve the closed-chain constraint is \emph{Projection} (\emph{PJ}). Several algorithms were developed based on it, such as $\emph{RGD}$ \cite{yakey2001randomized} and $\emph{CBiRRT}$ \cite{berenson2011task}. However, the constrained Jacobian is not guaranteed to be invertible all the time, and this approach relies heavily on the gradient descent operation, which is time-consuming for the high-dimensional \emph{MRS}. For example, \cite{zhang2021task} realized the motion planning of three \emph{MM}s in the simulated environment, but the mean planning time is over one minute. To improve the computational efficiency, \emph{Atlas} (\emph{AT}) \cite{jaillet2017path}, which uses piece-wise tangent spaces to locally approximate the constrained manifold, and \emph{Tangent Bundle} (\emph{TB}) \cite{kim2016tangent}, which is similar to \emph{AT} but with lazy states checking, were proposed. However, to get the tangent space, the kernel of the constrained Jacobian has to be computed, which requires complex matrix decomposition. Therefore, the computational efficiency of these framework will be reduced. 

In addition, to simplify the complex motion planning problem, several works try to solve the 3D transportion tasks into 2D formation control tasks \cite{oh2015survey}. For example, \cite{tang2018obstacle} defined a 2D virtual structure on the ground named system outlined rectangle ($\emph{SOR}$), and the leader-follower approach was applied to plan the motion of the system, where the leader adjusted the width of the $\emph{SOR}$ according to the environment. Similarly, the convex region was defined in \cite{alonso2017multi}, and then formation control was transformed into an optimization problem with respect to (w.r.t) the convex region. However, the obstacle-avoidance strategies in \cite{tang2018obstacle, alonso2017multi, jiao2015transportation} are conservative. As long as the virtual structure (e.g., \emph{SOR} or convex region) intersects with obstacles, the system is considered in collision. Therefore, bypassing becomes the only way to avoid obstacles, no matter how small they are (see Figure \ref{fig_sceneB_SOR}).

\begin{table*}[ht]
  \label{table_different_frameworks}
  \caption{Features of Different Motion Planning Frameworks When Dealing with Different Constraints}
  \begin{center}
  \footnotesize
  \setlength{\tabcolsep}{1mm}{
    \begin{tabular}{c|cccc}
      \hline
      Method                       & Closed-Chain  & Redundancy & Obstacle-Avoidance & Other Key Features                                                                                                                                                        \\ \hline
      Centralized Framework \cite{yakey2001randomized}\cite{berenson2011task}\cite{jaillet2017path}\cite{kim2016tangent}       & \checkmark                      & \ding{55}          & bypass and cross   & numerical projection; time-consuming   \\
      Virtual Structure-based Framework \cite{tang2018obstacle, alonso2017multi, jiao2015transportation}     & \checkmark                     & \ding{55}          & only bypass        & ---                 \\
      Fully Decoupled Hierarchical Framework \cite{hekmatfar2014cooperative}          & \checkmark                     & \ding{55}          & bypass and cross   & conflict between different layers         \\
      \textbf{Semi-Coupled Hierarchical Framework (\emph{ours})} & \checkmark                   & \checkmark         & bypass and cross   & \begin{tabular}[c]{@{}c@{}}time-efficient; compatibility between \\ different layers; 6D object's motion \end{tabular} \\ \hline
      \end{tabular}}
  \end{center}
\end{table*}

The above works (projection-based \cite{yakey2001randomized}\cite{berenson2011task}\cite{jaillet2017path}\cite{kim2016tangent} and virtual structure-based \cite{tang2018obstacle, alonso2017multi, jiao2015transportation} algorithms) plan the motion of the object and the robots at the same time, which can be seen as fully centralized methods. As the number of robots increases or the environment becomes cluttered, the computational complexity will increase and the success rate will decrease significantly (see Section \ref{section_motion_planner-overview}). Moreover, these methods are unable to deal with the redundancy of the \emph{MM} simultaneously and may cause discontinuous joint motion on the real robots. On the contrary, hierarchical or multilevel planning algorithm has shown its superiority in complex tasks \cite{orthey2020multilevel}. In this class of algorithms, special attention should be paid to the compatibility between adjacent layers (see Figure \ref{fig_poor_formation}a). For example, Hekmatfar et al. proposed a fully decoupled hierarchical framework to plan the motion of the multiple mobile manipulators system \cite{hekmatfar2014cooperative}. However, the object's motion planned by the centralized layer may not be executable by the robots in the decentralized layer due to obstacles or reachbility constraints, which will lead to failure to the cooperative transportation task. The features of different frameworks when dealing with the closed-chain, redundancy and obstacle-avoidance constraints are summarized in Table I.

\subsection{Contribution}

To deal with the closed-chain, redundancy and obstacle-avoidance constraints simultaneously, this paper proposes a motion planning framework with the following contributions:
\begin{itemize}
\item Semi-coupled hierarchical framework: the centralized layer plans the object's motion first, and then the redundancy of each \emph{MM} is explored independently in the decentralized layer. A notable feature is that closed-chain of the system, obstacle-avoidance of the object and the lower bound of the redundancy constraint metric are guaranteed while searching the path of the object in the centralized layer. This characteristic enables the decentralized layer to optimize each \emph{MM}'s joint motion while not violating object's motion and closed-chain constraint.
\item The framework can be applied to different numbers of \emph{MM}s and shows great superiority to the fully centralized planner and fully decoupled hierarchical planner, which are verified in the simulated experiments. 
\item To the best of our knowledge, this is the first time to achieve six-dimensional object transportation by multiple mobile manipulators in cluttered real-world environments.
\end{itemize}

The outline of this paper is as follows. Section \ref{section_problem_definition} will give a formal definition of the motion planning problem in detail. After that, we will introduce the proposed framework in Section \ref{section_motion_planner}. The feasibility and superiority of the framework will be verified by simulation and real-world experiments in Section \ref{section_exp}. Finally, the paper is concluded in Section \ref{section_conclusion}.

\section{PROBLEM DEFINITION} \label{section_problem_definition}

\subsection{Modeling of the Multiple Mobile Manipulators System}

Consider multiple mobile manipulators ($\emph{M}^3$) manipulating an object in Figure \ref{fig_robot_model}. Let $O_wX_wY_wZ_w$, $O_{obj}X_{obj}Y_{obj}Z_{obj}$, $O_{b,i}X_{b,i}Y_{b,i}Z_{b,i}$ and $O_{e,i}X_{e,i}Y_{e,i}Z_{e,i}$ be the frames of the world, the object, the mobile base and the end effector of the $i$th \emph{MM}, where $i = 1,2,...,n$ and $n$ is the number of the robots in the system. Some important notations are listed in Table II for convenience.

\begin{table}[ht]
  \caption{Definition of Some Important Notations}
  \label{important_notations}
  \centering
  \small 
  \setlength{\tabcolsep}{1mm}{
  \begin{tabular}{c|l}
    \hline
    $\boldsymbol{X}_i^j \in \text{SE}(3)$                                                                                               & Transformation of frame $i$ w.r.t frame $j$                                                     \\
    $\boldsymbol{t}_i^j = (\boldsymbol{p}_i^j{}^{\mathrm{T}}, \boldsymbol{\alpha}_i^j{}^{\mathrm{T}}){}^{\mathrm{T}} \in \mathbb{R}^6$  & Minimum representation of $\boldsymbol{X}_i^j$                                                  \\
    $\mathcal{C}_{\text{MM}}^i \subseteq \mathbb{R}^{n_{a,i} + n_{b,i}}$                                                                & Configuration space of the $i$th \emph{MM}                                                      \\
    $\boldsymbol{q}_{i} = (\boldsymbol{q}_{b,i}^T, \boldsymbol{q}_{a,i}^T)^T$                                                           & Joint configuration of the $i$th \emph{MM}                                                      \\
    $\mathcal{C}_{\text{obj}} \subseteq \mathbb{R}^6$                                                                                   & Configuration space of the object                                                               \\
    $\mathcal{C}_{\text{M}^3} \subseteq \mathbb{R}^{\sum_{i=1}^{n}(n_{a,i} + n_{b,i})+6}$                                               & Configuration space of the $M\textsuperscript{3}$ system                                        \\
    $\mathcal{C}_{\text{C}^3} \subseteq \mathcal{C}_{\text{M}^3}$                                                                       & Configuration space that satisfies $C\textsuperscript{3}$                                       \\
    $\boldsymbol{c} = (\boldsymbol{q}_1^T, \ldots, \boldsymbol{q}_n^T, \boldsymbol{t}_{\text{obj}}^w{}^T)^T$                            & Joint configuration of the $M\textsuperscript{3}$ system                                        \\
    $f_k(\cdot)$                                                                                                                        & Forward kinematic operator of the \emph{MM}                                                     \\
    $\boldsymbol{J}_i(\boldsymbol{q}_{i})$                                                                                              & Jacobian matrix of the $i$th \emph{MM}                                                          \\
    $\boldsymbol{E} = (\boldsymbol{t}_{e,1}^w{}^T, \ldots, \boldsymbol{t}_{e,n}^w{}^T)^T$                                               & End effector poses vector of each \emph{MM}                                                     \\
    $\boldsymbol{G} = (\boldsymbol{t}_{g,1}^w{}^T, \ldots, \boldsymbol{t}_{g,n}^w{}^T)^T$                                               & Grasping poses vector on the object                                                             \\
    $\pi: \mathcal{C}_{\text{M}^3} \rightarrow \mathcal{C}_{\text{obj}}$                                                                & Projection from $\mathcal{C}_{\text{M}^3}$ to $\mathcal{C}_{\text{obj}}$                          \\
    $g: \mathcal{C}_{\text{obj}} \rightarrow \mathbb{R}^{6n}$                                                                           & Mapping from $\mathcal{C}_{\text{obj}}$ to $\boldsymbol{G}$                                     \\
    $f_{\text{RC}}(\boldsymbol{c}): \mathcal{C}_{\text{C}^3} \rightarrow \mathbb{R}$                                                    & Redundancy constraint metric                                                                    \\
    $f_{\sss CC}(.)$                                                                                                                    & Compatibility condition about $f_{\text{RC}}$                                               \\
    $\Pi: \mathcal{C}_{\text{obj}} \rightarrow \mathcal{C}_{\text{C}^3}$                                                                & Mapping from $\mathcal{C}_{\text{obj}}$ to $\mathcal{C}_{\text{C}^3}$                           \\
    $\tau: [0,1] \rightarrow \mathcal{C}_{\text{C}^3}\cap \mathcal{C}_{\text{free}}$                                                    & A valid path of the $M\textsuperscript{3}$ system                                               \\
    \hline
  \end{tabular}}
\end{table}

The configurations of the $i$th mobile robot and manipulator are represented as $\boldsymbol{q}_{b,i}\in \mathbb{R}^{n_{b,i}}$ and $\boldsymbol{q}_{a,i}\in \mathbb{R}^{n_{a,i}}$, respectively. In most cases $n_{b,i}=3$ as the holonomic mobile robot can move and rotate in all directions on the ground. Therefore, the configuration space of the $i$th \emph{MM} can be defined as $\mathcal{C}_{\sss MM}^i \subseteq \mathbb{R}^{n_{a,i} + n_{b,i}}$, which is the set of configurations of the $i$th \emph{MM} $\boldsymbol{q}_{i} = (\boldsymbol{q}_{b,i}^T, \boldsymbol{q}_{a,i}^T)^T$. To represent the homogenous transformation of frame $i$ w.r.t frame $j$, we define $\boldsymbol{X}_i^j \in SE(3)$ and its minimum representation $\boldsymbol{t}_i^j = ({\boldsymbol{p}_i^j}^{\sss T},{\boldsymbol{\alpha}_i^j}^{\sss T})^T \in \mathbb{R}^6$, in which ${\boldsymbol{p}_i^j}$ denotes the relative position, and ${\boldsymbol{\alpha}_i^j}$ denotes a minimum description of orientation. Therefore, the configuration space of the object is defined as $\mathcal{C}_{obj} \subseteq \mathbb{R}^6$, which is the set of the object's poses with resect to the world frame $\boldsymbol{t}_{obj}^w$.

Let the configuration space of the system $\mathcal{C}_{\sss M^3} = \mathcal{C}_{\sss MM}^1 \times ... \times \mathcal{C}_{\sss MM}^i \times ... \times \mathcal{C}_{\sss MM}^n \times \mathcal{C}_{obj} \subseteq \mathbb{R}^{\sum_{i=1}^{n}{(n_{a,i} + n_{b,i})+6}}$ be the set of $\boldsymbol{c} = (\boldsymbol{q}_1^T, ..., \boldsymbol{q}_i^T, ..., \boldsymbol{q}_n^T, {\boldsymbol{t}_{obj}^w}^T)^T$. The forward kinematics of the $i$th \emph{MM} at position-level and velocity-level are defined as Eq. (\ref{eq_kin}) and Eq. (\ref{eq_diff_kin}), respectively.
\begin{equation}
\label{eq_kin}
    \boldsymbol{t}{_{e,i}^w} = f_k(\boldsymbol{q}_{i})
\end{equation}
\begin{equation}
\label{eq_diff_kin}
    \dot{\boldsymbol{t}}{_{e,i}^w} = \frac{\partial{f_k(\boldsymbol{q}_{i})}} {\partial{\boldsymbol{q}_{i}}} \dot{\boldsymbol{q}}_i
    = \boldsymbol{J}_i(\boldsymbol{q}_{i}) \dot{\boldsymbol{q}}_i
\end{equation}
where $f_k(.)$ is the forward kinematic operator, and $\boldsymbol{J}_i(\boldsymbol{q}_{i}) \in \mathbb{R}^{6 \times ({n_{a,i} + n_{b,i}})}$ is the analytical Jacobian matrix of the $i$th \emph{MM}. When a six-$\emph{DOF}$ (degree of freedom) manipulator mounting on an omnidirectional mobile robot, the column of $\boldsymbol{J}_i(\boldsymbol{q}_{i})$ is larger than that of the row, leading to redundancy of the \emph{MM}.

\subsection{Closed-Chain Constraint}\label{section_model_ccc}

In the proposed framew and the experiments, we make the follow assumptions.

$\bold{Assumption1:}$
\begin{itemize}
  \item \emph{the object is rigid and its geometric information is known;}
  \item \emph{the grasping pose w.r.t the object's pose is constant and known;}
  \item \emph{there is no slippage and deformation between the end effector and the object;}
  \item \emph{the model of the robots and the scene are known in advance;}
\end{itemize}

Given a random sample $\boldsymbol{c} = (\boldsymbol{q}_1^T, ..., \boldsymbol{q}_i^T, ..., \boldsymbol{q}_n^T, {\boldsymbol{t}_{obj}^w}^T)^T$, we define the vector of the end effector's poses as $\boldsymbol{E} = (\boldsymbol{t}{_{e,1}^w}^T,..., \boldsymbol{t}{_{e,i}^w}^T, ..., \boldsymbol{t}{_{e,n}^w}^T)^T \in \mathbb{R}^{6n}$ where $\boldsymbol{t}{_{e,i}^w} = f_k(\boldsymbol{q}_{i})$. The vector of the grasping poses on the object is denoted as $\boldsymbol{G} = (\boldsymbol{t}{_{g,1}^w}^T,..., \boldsymbol{t}{_{g,i}^w}^T, ..., \boldsymbol{t}{_{g,n}^w}^T)^T \in \mathbb{R}^{6n}$, in which $\boldsymbol{t}{_{g,i}^w}$ represents the $i$th grasping pose w.r.t the world frame. For convenience, a projection $\pi: \mathcal{C}_{\sss M^3} \rightarrow \mathcal{C}_{obj}$ is defined so that $\pi(\boldsymbol{c}) = \boldsymbol{t}_{obj}^w$. According to Assumption1, $\boldsymbol{G}$ can be easily derived by the geometric transformation $g: \mathcal{C}_{obj} \rightarrow \mathbb{R}^{6n}$.
\begin{equation}
\label{eq_grasp}
    \boldsymbol{G} = g(\pi(\boldsymbol{c}))
\end{equation}

When rigid grasp exists between $\boldsymbol{t}{_{e,i}^w}$ and $\boldsymbol{t}{_{g,i}^w}$ for all $i = 1,2,...,n$, the closed-chain constraint ($\emph{C}^3$) forms, and it can be formally described by $f_{\sss C^3}: \mathcal{C}_{\sss M^3} \rightarrow \mathbb{R}^{6n}$ in Eq. (\ref{eq_ccc1}).
\begin{equation}
\label{eq_ccc1}
 f_{\sss C^3}(\boldsymbol{c}) = \boldsymbol{E} - \boldsymbol{G} =\boldsymbol{0}
\end{equation}
Therefore, the set of configurations that satisfy the closed-chain constraint is denoted as $\mathcal{C}_{\sss C^3} \subseteq \mathcal{C}_{\sss M^3}$ in Eq. (\ref{eq_ccc2}).
\begin{equation}
\label{eq_ccc2}
    \mathcal{C}_{\sss C^3} = \{ \boldsymbol{c} | \boldsymbol{c} \in \mathcal{C}_{\sss M^3}, f_{\sss C^3}(\boldsymbol{c}) = \boldsymbol{0} \}
\end{equation}

\subsection{Redundancy Constraint}

Redundancy is a double-edged sword. On the one hand, it allows the \emph{MM} to keep desired configuration while following specified end effector path. On the other hand, it brings additional constraint to the motion planning problem. The fundamental difference between the closed-chain constraint and the redundancy constraint ($\emph{RC}$) is that $\emph{C}^3$ is ``hard'' but $\emph{RC}$ is ``soft'', in which ``hard'' constraints have to be satisfied exactly everywhere, and ``soft'' constraints are usually described by cost functions and optimized as much as possible. Therefore, $\emph{RC}$ can be transformed into an optimization problem and formally defined as follow.
\begin{equation}
\begin{aligned}
\label{eq_formation}
&\arg\max_{\boldsymbol{c}} \quad f_{\sss RC}(\boldsymbol{c})\\
&\begin{array}{r@{\quad}r@{}l@{\quad}l}
s.t. &\boldsymbol{c} \in \mathcal{C}_{\sss C^3}\cap \mathcal{C}_{free}
\end{array}
\end{aligned}
\end{equation}
where $\mathcal{C}_{free}$ represents the set of collision-free configurations with appropriate dimensions. $f_{\sss RC}: \mathcal{C}_{\sss C^3} \rightarrow \mathbb{R}$ is the normalized redundancy constraint metric of the system and will be introduced in Section \ref{section_motion_planner-decentralized}.

\subsection{Problem Definition}

In addition to the notations above, we define a mapping $\Pi: \mathcal{C}_{obj} \rightarrow \mathcal{C}_{\sss C^3}$ so that given a pose of the object $\boldsymbol{t}_{obj}^w$, $\Pi(\boldsymbol{t}_{obj}^w) = \{ \boldsymbol{c} | \boldsymbol{c} \in \mathcal{C}_{\sss C^3}\cap \mathcal{C}_{free}, \pi(\boldsymbol{c}) = \boldsymbol{t}_{obj}^w \}$. It is the set of configurations that satisfy collision and closed-chain constraints while the object's pose is invariant. Therefore, the motion planning of multiple mobile manipulators under closed-chain, redundancy and obstacle-avoidance constraints is defined as follows: given the world model $\mathcal{W}$, a start configuration $\boldsymbol{c}_{start} \in \mathcal{C}_{\sss C^3}\cap \mathcal{C}_{free}$ and a target pose of the object $\boldsymbol{t}_{obj}^w$, find a path $\tau: [0,1] \rightarrow \mathcal{C}_{\sss C^3}\cap \mathcal{C}_{free}$ such that 1) $\tau(0) = \boldsymbol{c}_{start}$; 2) $\tau(1) \in \Pi(\boldsymbol{t}_{obj}^w)$; and 3) $f_{\sss RC}(\boldsymbol{c})$ is maximized throughout $\tau$. The mathematical expression is shown in Eq. (\ref{eq_problem_definition1}).
\begin{equation}
  \label{eq_problem_definition1}
  \begin{array}{cl}
    \text { find } & \tau: [0,1] \rightarrow \mathcal{C}_{\sss C^3}\cap \mathcal{C}_{free} \subseteq \mathbb{R}^{6+\sum_{i=1}^{n}{(n_{a,i} + n_{b,i})}}  \\
    \text { s.t } & $\ding{172}$ \ \tau(0) = \boldsymbol{c}_{start} \\
                  & $\ding{173}$ \ \tau(1) \in \Pi(\boldsymbol{t}_{obj}^w) \\
                  & $\ding{174}$ \ \arg\max\limits_{\boldsymbol{c}} \ f_{\sss RC}(\boldsymbol{c}) 
  \end{array}
\end{equation}

\section{SEMI-COUPLED HIERARCHICAL MOTION PLANNING FRAMEWORK} \label{section_motion_planner}

\subsection{Overview of the Framework}\label{section_motion_planner-overview}

\begin{figure}[ht]
  \centering
  \includegraphics[width=7.5cm]{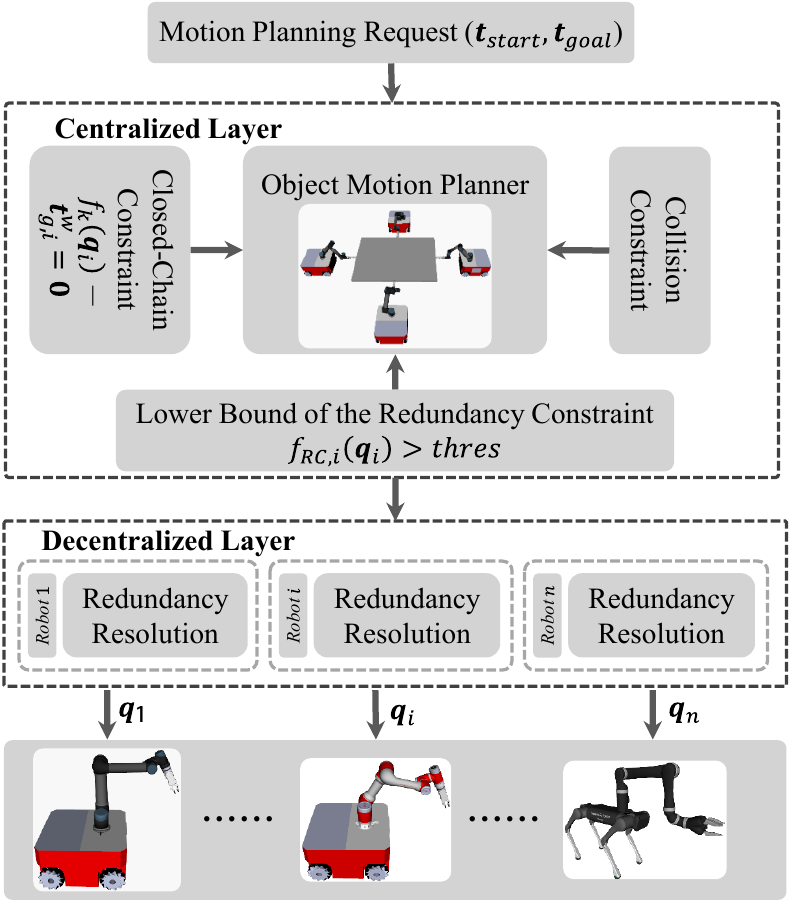}
  \caption{Overview of the semi-coupled hierarchical framework.}
  \label{fig_overview}
\end{figure}

One challenge of the motion planning problem is the high-dimensional configuration space. As discussed in Section \ref{section_related_work}, to satisfy the closed-chain constraint of the system, \emph{PJ}-like centralized frameworks will project a random state $\boldsymbol{c}_{rand} = (\boldsymbol{q}_1^T, ..., \boldsymbol{q}_i^T, ..., \boldsymbol{q}_n^T, {\boldsymbol{t}_{obj}^w}^T)^T \in \mathcal{C}_{\sss M^3}$ to a new state $\boldsymbol{c}_{new} \in \mathcal{C}_{\sss C^3}$ iteratively. Assuming there are $n$ robots and the object is required to achieve six-dimensional motion in the workspace, the size of the closed-chain constrained Jacobian $\boldsymbol{J}_{\sss C^3}(\boldsymbol{c})$ will be $6n \times \left(\sum_{i=1}^{n}{(n_{a,i} + n_{b,i})+6}\right)$ (see APPENDIX B). According to APPENDIX C, the time and space complexity when calculating the pseudo-inverse of $\boldsymbol{J}_{\sss C^3}(\boldsymbol{c})$ are $\mathcal{O}_t(n^3)$ and $\mathcal{O}_s(n^2)$, respectively. Therefore, when $n$ increases, the projection process (see APPENDIX A) will be extremely slow and easy to fall into local minimums.

Another challenge of solving Eq. (\ref{eq_problem_definition1}) is the coupling between closed-chain and redundancy constraints. Ideally, it is necessary to optimize the redundancy constraint metric on each sample $\boldsymbol{c}$ and generate a globally optimal path about $f_{\sss RC}(\boldsymbol{c})$. However, this is hard to achieve both technically and theoretically. On the one hand, most samples are not on the final path connecting $\tau(0)$ and $\tau(1)$, hence optimizing $f_{\sss RC}(\boldsymbol{c})$ on these samples is unnecessary and time-consuming. On the other hand, the convergence to optimality is not guaranteed for the existing optimal motion planners like $\emph{RRT}^*$ \cite{karaman2011sampling} when optimizing over something other than the path length.

Therefore, we decompose Eq. (\ref{eq_problem_definition1}) into the following two layers. Collision-free object's path is first planned in Eq. (\ref{eq_problem_definition2}), then the configuration of each robot is optimized in Eq. (\ref{eq_problem_definition3}).
\begin{equation}
  \label{eq_problem_definition2}
  \begin{array}{cl}
    \text { find } & \tau': [0,1] \rightarrow \mathcal{C}_{\sss obj} \cap \mathcal{C}_{free} \subseteq \mathbb{R}^{6} \\
    \text { s.t } & $\ding{172}$ \ \tau'(0) = \boldsymbol{t}_{start} \\
                  & $\ding{173}$ \ \tau'(1) = \boldsymbol{t}_{goal} \\
                  & \begin{cases}
                      $\ding{174}$ \ \exists \boldsymbol{q}_{i} \in \mathcal{C}_{free}, \ \ni f_k(\boldsymbol{q}_i)-\boldsymbol{t}_{g,i}^w = \boldsymbol{0} \\
                      $\ding{175}$ \ f_{\sss CC}(f_{\sss RC,i}(\boldsymbol{q}_i))
                    \end{cases}
  \end{array} 
\end{equation}
\begin{equation}
  \label{eq_problem_definition3}
  \begin{array}{cl}
     \arg\max\limits_{\boldsymbol{q}_i} & f_{\sss RC,i}(\boldsymbol{q}_i)  \\
     \text { s.t } & \boldsymbol{q}_i \in \mathbb{R}^{n_{a,i}+n_{b,i}} \cap \mathcal{C}_{free} \\
                  & f_k(\boldsymbol{q}_i)-\boldsymbol{t}_{g,i}^w = \boldsymbol{0}
  \end{array} 
\end{equation}

To alleviate the time cost brought by the high-dimensional configuration space, task space decomposition is applied to quickly check the closed-chain constraint. In condition \ding{174} of Eq. (\ref{eq_problem_definition2}), given the grasping pose $\boldsymbol{t}_{g,i}^w$, to meet the closed-chain constraint, each robot only need to project its own joint configuration $\boldsymbol{q}_i$ based on $\boldsymbol{J}_i(\boldsymbol{q}_{i})$ other than $\boldsymbol{J}_{\sss C^3}(\boldsymbol{c})$. According to Appendix C, the time and space complexity when calculating the pseudo-inverse of $\boldsymbol{J}_i(\boldsymbol{q}_{i}) (i=1,...,n)$ are $\mathcal{O}_t(n)$ and $\mathcal{O}_s(n)$, respectively. Therefore, the computation complexity will be reduced compared with \emph{PJ}-like centralized frameworks.

To alleviate the challenge brought by the coupling between closed-chain and redundancy constraints, the motion of the object and the robots are planned asynchronously. In addition, collision and closed-chain of the $i$th \emph{MM} (condition \ding{174}) and the compatibility condition (\emph{CC}) about $f_{\sss RC,i}(\boldsymbol{q}_i)$ (condition \ding{175}) are checked to avoid conflict between the two layers.

The framework composed by Eq. (\ref{eq_problem_definition2}) and Eq. (\ref{eq_problem_definition3}) is unified since different mapping $f_{\sss CC}(.)$ will lead to different motion planners. When condition \ding{174} exists and $f_{\sss CC}(.):=\arg\max_{\sss {\boldsymbol{q}_i}} f_{\sss RC,i}(\boldsymbol{q}_i)$, where $f_{{\sss RC},i}(\boldsymbol{q}_{i})$ is the normalized redundancy constraint metric of the $i$th \emph{MM}, Eq. (\ref{eq_problem_definition2}) will plan the object's moiton and optimize the joint configuration simultaneously, hence Eq. (\ref{eq_problem_definition3}) is no longer necessary, and this will lead to a fully centralized planner; When conditions \ding{174} and \ding{175} in Eq. (\ref{eq_problem_definition2}) do not exist, this is actually a fully decoupled framework, which is similar to \cite{hekmatfar2014cooperative}; When condition \ding{174} exists and $f_{\sss CC}(.):=f_{\sss RC,i}(\boldsymbol{q}_i) > \emph{thres}$, where $\emph{thres} \in [0,1]$ is a threshold of $f_{\sss RC,i}(\boldsymbol{q}_i)$, this will lead to a semi-coupled framework, which is shown in Figure \ref{fig_overview}.

In the semi-coupled framework, the ``soft'' redundancy constraint is relaxed in the first layer and delayed to the next layer for optimization. The parameter \emph{thres} can be seen as a coupling factor. As it approaches maximum value 1, the coupling between the two layers gradually intensifies, which will bring more computational burden to Eq. (\ref{eq_problem_definition2}). As it approaches minimum value 0, the two layers gradually decouple, which may cause conflict between them. Therefore, \emph{thres} should be carefully selected so that the compatibility between the two layers is ensured while not putting much computation burden to the former. The detail of the two layers is introduced in the following.

\subsection{Centralized Layer --- Object Motion Planner}\label{section_motion_planner-cenlayer}

\IncMargin{1em}
\begin{algorithm}
\caption{\emph{Object Motion Planner}}\label{alg1}
  \SetKwProg{Fn}{Function}{}{}
  \Fn{\texttt{\emph{plan}}($\boldsymbol{t}_{start}$, $\boldsymbol{t}_{goal}$, $t_{obj}$, thres, $\mathcal{W}$)}
  {
    \If(){\texttt{\emph{validChecking}}($\boldsymbol{t}_{goal}$, thres) \emph{is false}}
    {
        \KwRet fail\;
    }
    $\mathcal{T}_s.\texttt{init}(\boldsymbol{t}_{start})$\;
    $\mathcal{T}_g.\texttt{init}(\boldsymbol{t}_{goal})$\;
    \For{$t\leftarrow 0$ \KwTo $t_{obj}$}
    {
        $\boldsymbol{t}_{rand}$ $\leftarrow$ \texttt{randomSample}()\;
        $\boldsymbol{t}_{near}$ $\leftarrow$ \texttt{nearestSearch}($\mathcal{T}_s$, $\boldsymbol{t}_{rand}$)\;
        $\boldsymbol{t}_{new}$ $\leftarrow$ \texttt{extend}($\boldsymbol{t}_{near}$, $\boldsymbol{t}_{rand}$)\;
        \If(){\texttt{\emph{validChecking}}($\boldsymbol{t}_{new}$,thres) \emph{is true}}
        {
          $\mathcal{T}_s.\texttt{add}(\boldsymbol{t}_{near}, \boldsymbol{t}_{new})$\;
        }
        \If(){\texttt{\emph{connect}}($\mathcal{T}_s, \mathcal{T}_g$) \emph{is success}}
        {
            \KwRet \texttt{extractPath}($\mathcal{T}_s, \mathcal{T}_g$)\;
        }
        \texttt{swap}($\mathcal{T}_s, \mathcal{T}_g$)\;
    }
    \KwRet fail\;
  }

  \Fn{\texttt{\emph{validChecking}}($\boldsymbol{t}_{rand}$, thres)}
  {
    \If{$\boldsymbol{t}_{rand}$ \emph{is in collision}}
    {
        \KwRet false\;
    }
    \tcp{compute G according to Eq.(\ref{eq_grasp})}
    $\boldsymbol{G}$ $\leftarrow$ \texttt{getGraspingPoses}($\boldsymbol{t}_{rand}$)\;
    \For{\textbf{\emph{each}} $\boldsymbol{t}_{g,i}^w$ \textbf{\emph{in}} $\boldsymbol{G}$}
    {
        \tcp{defined in Alg.(\ref{alg_isASREmpty})}
        \If{\texttt{\emph{isASREmpty}}($\boldsymbol{t}_{g,i}^w$, thres)}
        {
            \KwRet false\;
        }
    }
    \KwRet true\;
  }
\end{algorithm}\DecMargin{1em}

Alg. \ref{alg1} shows the object's motion planner in the structure of $\emph{RRTConnect}$ algorithm \cite{kuffner2000rrt}. By encapsulating the validity checking module, which checks collision, closed-chain and lower bound of the redundancy constraint metric, into an independent component, working with other sampling-based algorithms, such as $\emph{PRM}$ \cite{kavraki1996probabilistic} and $\emph{EST}$ \cite{hsu1997path} is also possible. For brevity of notations, we omit the superscript and subscript of $\boldsymbol{t}_{obj}^w$ in this part. For example, $\boldsymbol{t}_{start}$, $\boldsymbol{t}_{goal}$ and $\boldsymbol{t}_{rand}$ represent the start, the goal and random pose of the object w.r.t the world frame, separately. The key functions in Alg. \ref{alg1} are explained as follows.

\begin{itemize}[leftmargin=*]
\item $\texttt{plan}(\boldsymbol{t}_{start}, \boldsymbol{t}_{goal}, t_{obj}, \emph{thres}, \mathcal{W})$ generates collision-free object's path when the planning request ($\boldsymbol{t}_{start}, \boldsymbol{t}_{goal}$), allowed planning time $t_{obj}$ and the world model $\mathcal{W}$ are specified. Before planning, the validity of $\boldsymbol{t}_{goal}$ is first checked, and two trees are initialized, where $\mathcal{T}_s$ starts from $\boldsymbol{t}_{start}$ and $\mathcal{T}_g$ starts from $\boldsymbol{t}_{goal}$. Then a random object's pose $\boldsymbol{t}_{rand}$ and its nearest neighbor $\boldsymbol{t}_{near}$ on $\mathcal{T}_s$ are computed. After that, a new node $\boldsymbol{t}_{new}$ is generated from $\boldsymbol{t}_{near}$ to $\boldsymbol{t}_{rand}$ along the given step size. When $\boldsymbol{t}_{new}$ is valid, the edge connecting $\boldsymbol{t}_{near}$ and $\boldsymbol{t}_{new}$ will be added to $\mathcal{T}_s$. In each iteration, the two trees try to connect with each other and extract the valid path. If the two trees are unable to connect, they are swapped and continue to grow in the next iteration until the path is found or the allowed time $t_{obj}$ is exceeded.

\item $\texttt{validChecking}(\boldsymbol{t}_{rand}, \emph{thres})$ checks whether $\boldsymbol{t}_{rand}$ can be extended to a composite configuration $\boldsymbol{c}_{rand} \in \mathcal{C}_{\sss C^3}\cap \mathcal{C}_{free}$ so that $\pi(\boldsymbol{c}_{rand}) = \boldsymbol{t}_{rand}$ and $f_{\sss RC}(\boldsymbol{c})$ is larger than \emph{thres}. When the object is collision-free, the grasping poses vector $\boldsymbol{G}$ will be calculated based on Eq. (\ref{eq_grasp}). For each pose $\boldsymbol{t}_{g,i}^w$ in $\boldsymbol{G}$, the allowed sampling region is sampled to check whether the closed-chain, collision and the user-specified lower bound of the redundancy constraint metric can be satisfied for each \emph{MM}. When these conditions are met for all $i = 1,2,...,n$, $\boldsymbol{t}_{rand}$ is seen as a valid sample. The detail of $\texttt{isASREmpty()}$ is introduced in the following.
\end{itemize}

\begin{figure}[ht]
  \centering
  \includegraphics[width=7.5cm]{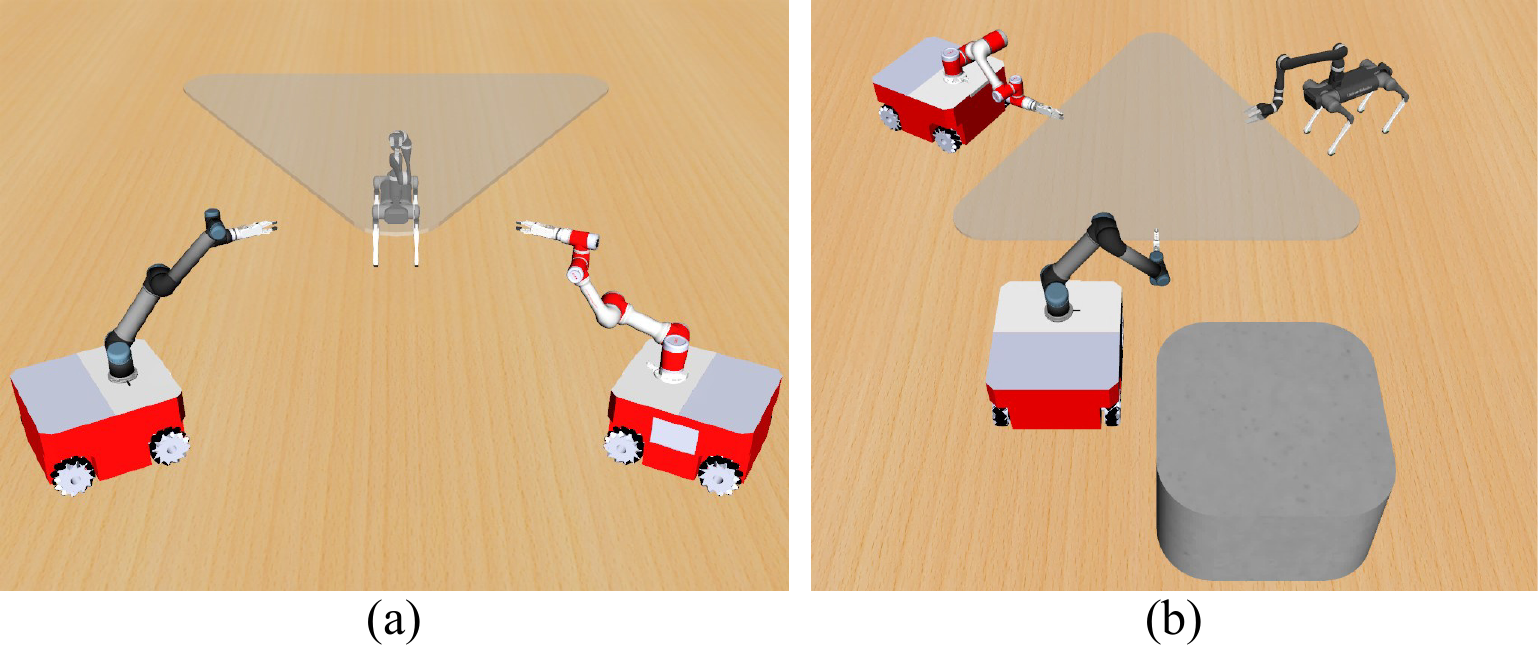}
  \caption{Confliction between the centralized layer and the decentralized layer. }
  \label{fig_poor_formation}
\end{figure}

\emph{Allowed Sampling Region}: In the hierarchical framework, the compatibility between adjacent layers should be carefully considered. Otherwise, the centralized layer may conflict with the decentralized layer. For example, although the object's pose is collision-free in Figure \ref{fig_poor_formation}(a), it exceeds the workspace of the robots, and the closed-chain constraint is unable to be satisfied. Similarly, for the object's pose in Figure \ref{fig_poor_formation}(b), although a configuration can be found to satisfy the closed-chain constraint, one of the robots must be in an awkward configuration due to obstacles. Therefore, the object's poses in these cases should be abandoned as they are inexecutable or lead to poor and unstable joint configuration in the decentralized layer.

To check the validity of $\boldsymbol{t}_{g,i}^w$, we define the allowed sampling region ($\emph{ASR}$) for each mobile robot in Eq. (\ref{eq_asr}).
\begin{equation}
\label{eq_asr}
\begin{split}
    ASR(\boldsymbol{t}_{g,i}^w, & \emph{thres}) = \{ \boldsymbol{q}_{b,i} \ | \ \exists \boldsymbol{q}_{a,i}, \ni f_k(\boldsymbol{q}_{i}) = \boldsymbol{t}_{g,i}^w \\
     &  \ \emph{and} \ f_{{\sss RC},i}(\boldsymbol{q}_{i}) \geq \emph{thres} \ \emph{and} \ \boldsymbol{q}_{i} \in \mathcal{C}_{free} \}
\end{split}
\end{equation}
where $f_{{\sss RC},i}(\boldsymbol{q}_{i})$ is the normalized redundancy constraint metric of the $i$th \emph{MM}, and $\emph{thres}$ is a user-specified threshold of this metric. $f_{{\sss RC},i}(\boldsymbol{q}_{i})$ is a component of $f_{\sss RC}(\boldsymbol{c})$, and their expressions are defined in Eq. (\ref{eq_redundancy1}) and Eq. (\ref{eq_redundancy2}). Therefore, we can check whether existing $\boldsymbol{q}_{b,i} \in ASR(\boldsymbol{t}_{g,i}^w, \emph{thres})$ to see the validity of $\boldsymbol{t}_{g,i}^w$.

\IncMargin{1em}
\begin{algorithm}
\caption{\texttt{isASREmpty}($\boldsymbol{t}_{g,i}^w$, $\emph{thres}$)}\label{alg_isASREmpty}
    $\emph{bounds}$ $\leftarrow$ \texttt{getMobileRobotBounds}($\boldsymbol{t}_{g,i}^w$)\;
    \For{$j\leftarrow 1$ \KwTo maxSample}
    {
        $\boldsymbol{q}_{b,i}$ $\leftarrow$ \texttt{sampleMobileRobot}($\emph{bounds}$)\;
        $\boldsymbol{q}_{i}$ $\leftarrow$   \texttt{computeIK}($\boldsymbol{t}_{g,i}^w$, $\boldsymbol{q}_{b,i}$)\;
        $f_{{\sss RC},i}$ $\leftarrow$      \texttt{computeRedundancyMetric}($\boldsymbol{q}_{i}$)\;
        \If{\texttt{\emph{collisionFree}}($\boldsymbol{q}_{i}$) \emph{and} $f_{{\sss RC},i} >$ thres}
        {
          \KwRet false\;
        }
    }
    \KwRet true\;
\end{algorithm}\DecMargin{1em}

$\emph{ASR}$ changes with $\boldsymbol{t}_{g,i}^w$ and $\emph{thres}$. Due to the complexity of the system, it is usually obtained by sampling. In cooperative transportation tasks, the mobile robot has almost unlimited range of motion on the ground but the manipulator only has limitted reachable workspace, which can be seen as heuristic information. In Alg. \ref{alg_isASREmpty}, given the grasping pose $\boldsymbol{t}_{g,i}^w$, we can get the approximate motion \emph{bounds} for the mobile robot based on the maximum workspace of the manipulator, and this idea has been applied for the \emph{MM} in the literatures \cite{zhang2020novel}\cite{thakar2022manipulator}. Then given a random $\boldsymbol{q}_{b,i}$ within \emph{bounds}, the joint configuration of the \emph{MM} $\boldsymbol{q}_i$ can be computed based on the iterative inverse kinematic solvers. Finally, if $\boldsymbol{q}_i$ is not in collision and the corresponding redundancy constraint metric $f_{{\sss RC},i}(\boldsymbol{q}_{i})$ is larger than \emph{thres}, the set $ASR(\boldsymbol{t}_{g,i}^w, \emph{thres})$ is not empty and $\boldsymbol{t}_{g,i}^w$ can be seen as a valid grasping pose.

\subsection{Decentralized Layer --- Redundancy Resolution}\label{section_motion_planner-decentralized}

As shown in Figure \ref{fig_overview} and Eq. (\ref{eq_problem_definition2})-(\ref{eq_problem_definition3}), redundancy resolution is an independent component in the semi-coupled hierarchical framework, and various off-the-shelf algorithms can be used in the decentralized layer. For example, our previous work \cite{zhang2022cooperative} proposed a real-time and robust redundancy resolution for the \emph{MM} in the human-robot cooperative transportation task, hence we will reuse this algorithm and briefly introduce it in the following. However, as long as the redundancy constraint is modeled as an optimization problem, such as \cite{ancona2017redundancy}, it can be combined with the proposed framework.

The key to the redundancy resolution is fulfilling additional constraints while obeying the end effector's motion, hence the following constraints are introduced for the $i$th \emph{MM}.
\begin{equation}
  \label{eq_redundancy1}
    f_{{\sss RC},i}(\boldsymbol{q}_i) = 
      \mu(\boldsymbol{q}_i) \times F(\boldsymbol{q}_i) \times B(\boldsymbol{q}_i) 
\end{equation}
where $\mu(\boldsymbol{q}_i)$, $F(\boldsymbol{q}_i)$ and $B(\boldsymbol{q}_i)$ represent the dexterity of the \emph{MM}, the formation metric of the \emph{MRS} and obstacle-avoidance constraint index of the mobile robot, respectively. These indexes are relative values w.r.t their global optimum, hence are normalized within [0,1].

When they are small, the planner will generate a better configuration in the next round of optimization. For example, when the dynamic obstacles approach the mobile robot, $B(\boldsymbol{q}_i)$ will be punished, and the mobile robot will move far away from the obstacles in the next iteration. An intuitive explanation of Eq. (\ref{eq_redundancy1}) is as follows. As a redundant robot, there are extra \emph{DOF} to fulfill the user-specified constraints. In cooperative transportation task, the \emph{MM} should adjust self-motion to avoid obstacles on the ground and keep a dexterous configuration and optimize the formation of the system as much as possible. For more detail, please refer to \cite{zhang2022cooperative}.

Therefore, the redundancy constraint metric of the system $f_{\sss RC}(\boldsymbol{c})$ in Eq. (\ref{eq_formation}) can be defined as the minimum $f_{{\sss RC},i}(\boldsymbol{q}_i)$ of all robots. When $f_{{\sss RC},i}(\boldsymbol{q}_i)$ reaches its maximum for all robots, the redundancy of the system is optimized.
\begin{equation}
\label{eq_redundancy2}
 f_{\sss RC}(\boldsymbol{c}) = \emph{min}\{f_{{\sss RC},i}(\boldsymbol{q}_i) | 1 \leq i \leq n \}
\end{equation}
In the simulation and real-world experiments, Nelder-Mead simplex algorithm \cite{nelder1965simplex} is chosen as the real-time solver to get optimal joint configuration $\boldsymbol{q}_i$ about $f_{{\sss RC},i}(\boldsymbol{q}_i)$. It has no requirement on the differentiability of $f_{{\sss RC},i}$ and shows good performance in practice.

\section{SIMULATION AND EXPERIMENTAL RESULTS} \label{section_exp}

\subsection{Experimental Setup}

To demonstrate the superiority of the proposed framework, numerous experiments are conducted in Section \ref{section_exp_sim} and Section \ref{section_exp_real}, respectively. Moreover, different numbers of robots are considered in the experiments. All the codes are implemented in the architecture of $\emph{ROS}$ based on $\emph{C++}$. The computer is Intel i7-8700 (3.2GHz) with 32GB memory.

According to Assumption1, $f_k(\boldsymbol{q}_i)$ in Eq. (\ref{eq_kin}) and $g(\boldsymbol{t}_{obj}^w)$ in Eq. (\ref{eq_grasp}) can be derived from the geometric model of the \emph{MRS}. $f_{\sss CC}(.)$ in Eq. (\ref{eq_problem_definition2}) is set to $f_{\sss RC,i}(\boldsymbol{q}_i) > \emph{thres}$ to couple the two layers. The convergence criteria for the projection process is set to 0.005 in \emph{SCHF}, which works well in the experiment. The expression of $f_{\sss RC,i}(\boldsymbol{q}_i)$ is shown in Eq. (\ref{eq_redundancy1}) and the suitable value of \emph{thres} will be tested and discussed in Section \ref{section_exp_sim4}.

\subsection{Performance of the Semi-Coupled Hierarchical Framework} \label{section_exp_sim}

\subsubsection{Simulation Scene}

As obstacles or the number of robots increase, the scenes are shown in Figure \ref{fig_sim_scene}(a)-(f). The size of the scenes is $15m \times 25m$. The start and goal states are displayed on the most left and most right, respectively. Snapshots during the tasks are also shown in Figure \ref{fig_sim_scene}. The pictorial motion is available in the attached video. In these tasks, the start and goal states are assumed to satisfy the closed-chain and collision constraints, and valid paths connecting the start and goal states exist theoretically. Moreover, the height of some obstacles is lower than the maximum allowed height of the object, which means that obstacles may be crossed by the system.

\subsubsection{Simulation 1} \label{section_exp_sim1}

\begin{figure*}[ht]
  \centering
  \includegraphics[width=14cm]{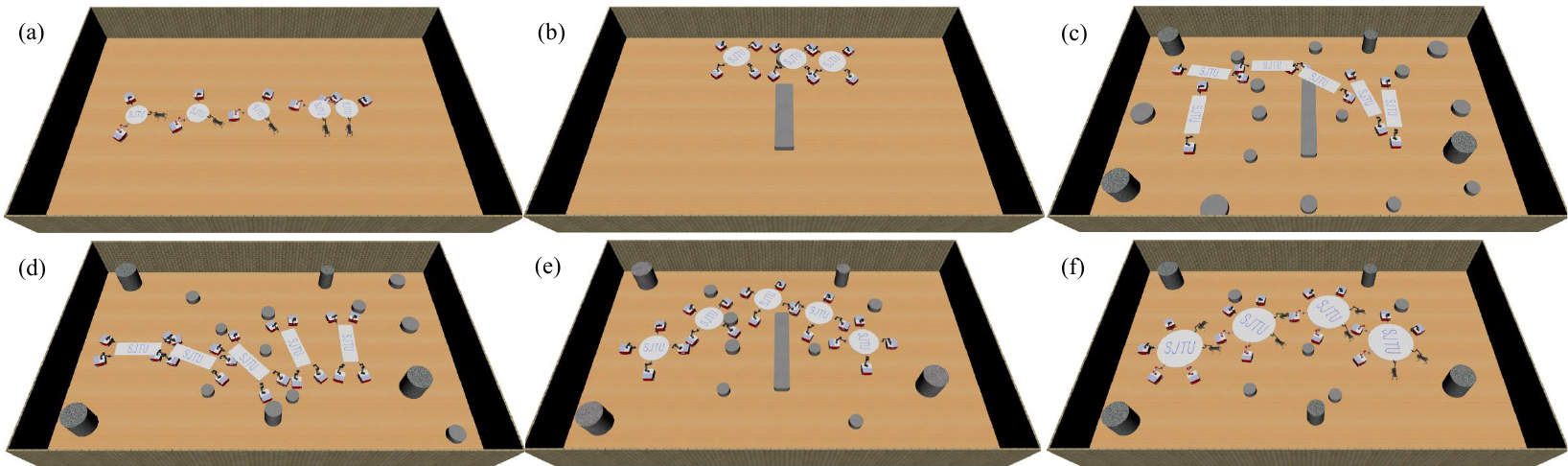}
  \caption{(a)-(f). Simulation scenes from simple to complex. The size of the scenes is $15m \times 25m$. Robot models are snapshots during the tasks. The height of some obstacles is lower than the maximum allowed height of the object, hence obstacles may be crossed by the system like the path in (b)-(f).}
  \label{fig_sim_scene}
\end{figure*}

\begin{table*}[ht]
\label{table_planner}
\caption{Comparison of the Semi-coupled Hierarchical Framework (\emph{SCHF}) with Centralized Frameworks (\emph{PJ} \cite{yakey2001randomized}\cite{berenson2011task}, \emph{AT} \cite{jaillet2017path} and \emph{TB} \cite{kim2016tangent}) }
\begin{center}
\scriptsize
\resizebox{\linewidth}{!}{
\begin{tabular}{c|l|cccccccccccc}
\hline
\multirow{2}{*}{Task}    & \multicolumn{1}{c|}{\multirow{2}{*}{Mehtod}}                     & \multicolumn{2}{c|}{$\emph{RRT}$\cite{lavalle1998rapidly}}                                                                               & \multicolumn{2}{c|}{$\emph{RRTConnect}$\cite{kuffner2000rrt}}                                                                            & \multicolumn{2}{c|}{$\emph{BKPIECE}$\cite{csucan2009kinodynamic}}                                                                        & \multicolumn{2}{c|}{$\emph{EST}$\cite{hsu1997path}}                                              & \multicolumn{2}{c|}{$\emph{STRIDE}$\cite{gipson2013resolution}}                                                                          & \multicolumn{2}{c}{$\emph{PRM}$\cite{kavraki1996probabilistic}}                                           \\ \cline{3-14}
                         & \multicolumn{1}{c|}{}                                            & \begin{tabular}[c]{@{}c@{}}Success/\\ Total\end{tabular} & \multicolumn{1}{c|}{\begin{tabular}[c]{@{}c@{}} Time(s)\end{tabular}} & \begin{tabular}[c]{@{}c@{}}Success/\\ Total\end{tabular} & \multicolumn{1}{c|}{\begin{tabular}[c]{@{}c@{}}Time(s)\end{tabular}} & \begin{tabular}[c]{@{}c@{}}Success/\\ Total\end{tabular} & \multicolumn{1}{c|}{\begin{tabular}[c]{@{}c@{}}Time(s)\end{tabular}} & \begin{tabular}[c]{@{}c@{}}Success/\\ Total\end{tabular} & \multicolumn{1}{c|}{\begin{tabular}[c]{@{}c@{}}Time(s)\end{tabular}} & \begin{tabular}[c]{@{}c@{}}Success/\\ Total\end{tabular} & \multicolumn{1}{c|}{\begin{tabular}[c]{@{}c@{}}Time(s)\end{tabular}} & \begin{tabular}[c]{@{}c@{}}Success/\\ Total\end{tabular} & \begin{tabular}[c]{@{}c@{}}Time(s)\end{tabular} \\ \hline
\multirow{5}{*}{Figure \ref{fig_sim_scene}(a)}  & \emph{PJ}\cite{yakey2001randomized}\cite{berenson2011task}     & 0/10           & ---                    & 0/10           & ---                    & 0/10          & ---                      & 9/10  & 7.75$\pm$6.39   & 2/10  & 11.87$\pm$3.80  & 10/10 & 23.48$\pm$2.34                                                      \\
                                            & \emph{AT}\cite{jaillet2017path}                                & 0/10           & ---                    & 10/10          & 3.37$\pm$0.55          & 0/10          & ---                      & 10/10 & 1.78$\pm$1.11   & 7/10  & 6.59$\pm$6.28   & 10/10 & 21.73$\pm$1.53                                                      \\
                                            & \emph{TB}\cite{kim2016tangent}                                 & 3/10           & 17.64$\pm$3.80         & 10/10          & 2.29$\pm$0.43          & 0/10          & ---                      & 10/10 & 1.58$\pm$1.71   & 7/10  & 3.60$\pm$2.72   & 10/10 & 22.31$\pm$1.55                                                      \\
                                            & $\emph{\textbf{SCHF (ours)}}$                                  & 10/10          & 7.34$\pm$4.34          & \textbf{10/10}          & \textbf{0.97$\pm$0.10}          & 0/10          & ---                      & 7/10  & 5.20$\pm$3.62   & 4/10  & 3.60$\pm$3.70   & 10/10 & 20.25$\pm$0.19                                                      \\ \hline

\multirow{5}{*}{Figure \ref{fig_sim_scene}(b)}  & \emph{PJ}\cite{yakey2001randomized}\cite{berenson2011task}     & 0/10           & ---                    & 0/10           & ---                    & 0/10          & ---                      & 0/10  & ---             & 0/10  & ---             & 0/10  & ---                                                      \\
                                            & \emph{AT}\cite{jaillet2017path}                                & 4/10           & 12.85$\pm$4.92         & 8/10           & 4.21$\pm$2.76          & 0/10          & ---                      & 5/10  & 6.12$\pm$6.49   & 0/10  & ---             & 6/10  & 22.90$\pm$1.81                                                      \\
                                            & \emph{TB}\cite{kim2016tangent}                                 & 7/10           & 10.64$\pm$2.40         & 8/10           & 5.71$\pm$5.75          & 0/10          & ---                      & 5/10  & 4.67$\pm$3.81   & 0/10  & ---             & 9/10  & 23.75$\pm$6.42                                                      \\
                                            & $\emph{\textbf{SCHF (ours)}}$                                  & \textbf{10/10}          & \textbf{0.81$\pm$1.11}          & 10/10          & 3.34$\pm$1.56          & 3/10          & 20.01$\pm$2.56           & 10/10 & 12.17$\pm$6.05  & 10/10 & 8.77$\pm$6.56   & 8/10  & 20.20$\pm$0.18                                                      \\ \hline

\multirow{5}{*}{Figure \ref{fig_sim_scene}(c)}  & \emph{PJ}\cite{yakey2001randomized}\cite{berenson2011task}     & 0/10           & ---                    & 0/10           & ---                    & 0/10          & ---                      & 0/10  & ---             & 0/10  & ---             & 0/10  & ---                                                      \\
                                            & \emph{AT}\cite{jaillet2017path}                                & 0/10           & ---                    & 0/10           & ---                    & 0/10          & ---                      & 0/10  & ---             & 0/10  & ---             & 0/10  & ---                                                                                                            \\
                                            & \emph{TB}\cite{kim2016tangent}                                 & 0/10           & ---                    & 0/10           & ---                    & 0/10          & ---                      & 0/10  & ---             & 0/10  & ---             & 0/10  & ---                                                                                                           \\
                                            & $\emph{\textbf{SCHF (ours)}}$                                  & 10/10          & 8.82$\pm$6.09          & \textbf{10/10}          & \textbf{6.99$\pm$5.71}          & 0/10          & ---                      & 8/10  & 14.40$\pm$8.57  & 9/10  & 17.77$\pm$8.37  & 7/10  & 35.32$\pm$5.72                                                      \\ \hline

\multirow{5}{*}{Figure \ref{fig_sim_scene}(d)}  & \emph{PJ}\cite{yakey2001randomized}\cite{berenson2011task}     & 0/10           & ---                    & 0/10           & ---                    & 0/10          & ---                      & 0/10  & ---             & 0/10  & ---             & 0/10  & ---                                                      \\
                                            & \emph{AT}\cite{jaillet2017path}                                & 0/10           & ---                    & 0/10           & ---                    & 0/10          & ---                      & 0/10  & ---             & 0/10  & ---             & 0/10  & ---                                                      \\
                                            & \emph{TB}\cite{kim2016tangent}                                 & 0/10           & ---                    & 0/10           & ---                    & 0/10          & ---                      & 0/10  & ---             & 0/10  & ---             & 0/10  & ---                                                      \\
                                            & $\emph{\textbf{SCHF (ours)}}$                                    & 10/10          & 8.30$\pm$4.48          & \textbf{10/10}          & \textbf{7.98$\pm$7.06}          & 0/10          & ---                      & 7/10  & 16.79$\pm$9.34  & 4/10  & 22.46$\pm$2.99  & 10/10 & 33.47$\pm$1.10                                                      \\ \hline

\multirow{5}{*}{Figure \ref{fig_sim_scene}(e)}  & \emph{PJ}\cite{yakey2001randomized}\cite{berenson2011task}     & 0/10           & ---                    & 0/10           & ---                    & 0/10          & ---                      & 0/10  & ---             & 0/10  & ---             & 0/10  & ---                                                      \\
                                            & \emph{AT}\cite{jaillet2017path}                                & 0/10           & ---                    & 0/10           & ---                    & 0/10          & ---                      & 0/10  & ---             & 0/10  & ---             & 0/10  & ---                                                      \\
                                            & \emph{TB}\cite{kim2016tangent}                                 & 0/10           & ---                    & 0/10           & ---                    & 0/10          & ---                      & 0/10  & ---             & 0/10  & ---             & 0/10  & ---                                                      \\
                                            & $\emph{\textbf{SCHF (ours)}}$                                  & \textbf{10/10}          & \textbf{23.84$\pm$12.28}        & 10/10          & 27.09$\pm$12.72        & 0/10          & ---                      & 0/10  & ---             & 1/10  & 46.41$\pm$0     & 6/10  & 54.41$\pm$2.66                                                      \\ \hline

\multirow{5}{*}{Figure \ref{fig_sim_scene}(f)}  & \emph{PJ}\cite{yakey2001randomized}\cite{berenson2011task}     & 0/10           & ---                    & 0/10           & ---                    & 0/10          & ---                      & 0/10  & ---             & 0/10  & ---             & 0/10  & ---                                                      \\
                                            & \emph{AT}\cite{jaillet2017path}                                & 0/10           & ---                    & 0/10           & ---                    & 0/10          & ---                      & 0/10  & ---             & 0/10  & ---             & 0/10  & ---                                                      \\
                                            & \emph{TB}\cite{kim2016tangent}                                 & 0/10           & ---                    & 0/10           & ---                    & 0/10          & ---                      & 0/10  & ---             & 0/10  & ---             & 0/10  & ---                                                      \\
                                            & $\emph{\textbf{SCHF (ours)}}$                                  & 10/10          & 10.17$\pm$3.89         & \textbf{10/10}          & \textbf{5.73$\pm$1.91}          & 1/10          & 47.58$\pm$0              & 10/10 & 26.06$\pm$10.89 & 10/10 & 19.10$\pm$6.51  & 10/10 & 53.95$\pm$1.35                                                      \\ \hline
\end{tabular}}
\end{center}
\end{table*}

In this simulation, we compare the proposed framework with centralized frameworks, such as \emph{PJ} \cite{yakey2001randomized}\cite{berenson2011task}, \emph{AT} \cite{jaillet2017path} and \emph{TB} \cite{kim2016tangent}. As discussed in Section \ref{section_related_work}, \emph{PJ} computes the Jacobian of the system and projects a random sample $\boldsymbol{c}$ onto the manifold based on Newton procedure (see APPENDIX). \emph{AT} and \emph{TB} are built on top of \emph{PJ} with some modifications. For the fairness of the comparison, in the semi-coupled hierarchical framework (\emph{SCHF}), the inverse kinematic solver is also chosen as an iterative algorithm.

In the tasks shown in Figure \ref{fig_sim_scene}(a)-(f), the start state $\boldsymbol{t}_{start}$ and goal state $\boldsymbol{t}_{goal}$ of the object are specified by the user. However, in \emph{PJ}, \emph{AT} and \emph{TB}, not only the state of the object but also the state of the robots should be specified. Therefore, a random goal state $\boldsymbol{c}_{goal}$ will be selected for each scene so that $\boldsymbol{c}_{goal} \in \mathcal{C}_{\sss C^3}\cap \mathcal{C}_{free}$ and $\pi(\boldsymbol{c}_{goal}) = \boldsymbol{t}_{goal}$. Moreover, redundancy can not be resolved directly in \emph{PJ}, \emph{AT} and \emph{TB}, hence only closed-chain and collision constraints are considered in these centralized frameworks. \emph{PJ}, \emph{AT} and \emph{TB} on the multiple mobile manipulators system are developed based on $\emph{OMPL}$ \cite{sucan2012open}, which is a famous sampling-based motion planning software.

The performances of different frameworks are shown in Table III. The terms and settings are explained as follows. The maximum allowed planning time $t_{obj}$ in Alg. \ref{alg1} is set to $20s$, $20s$, $30s$, $30s$, $50s$ and $50s$ for the tasks in Figure \ref{fig_sim_scene}(a)-(f), respectively. \emph{maxSample} in Alg. \ref{alg_isASREmpty} is set to 200, and $\emph{thres}$ is set to 0.4 to couple the centralized and decentralized layers. Each framework is combined with six random planning algorithms, namely $\emph{RRT}$ \cite{lavalle1998rapidly}, $\emph{RRTConnect}$ \cite{kuffner2000rrt}, $\emph{BKPIECE}$ \cite{csucan2009kinodynamic}, $\emph{EST}$ \cite{hsu1997path}, $\emph{STRIDE}$\cite{gipson2013resolution} and $\emph{PRM} $\cite{kavraki1996probabilistic}. Please notice that sampling-based motion planner like $\emph{RRT}$ is a sub-module of the proposed framework and only used to search the object's motion other than the \emph{MM}'s joint moiton. Each combination runs 10 times to get the statistical results. ``Success/Total" columns represent the successful and total simulations conducted. ``Time(s)" columns represent the mean time and the standard deviation of the successful simulations. When all simulations fail, ``Time(s)" columns are set to ``---".

From Figure \ref{fig_sim_scene} and Table III, we can see that the performance of \emph{PJ}, \emph{AT} and \emph{TB} increases sequentially on these tasks. They work well when the scenes are simple, like the task in Figure \ref{fig_sim_scene}(a). When obstacles increase, their performance degrades significantly. Moreover, we also find that \emph{PJ}-like frameworks are sensitive to the hyper-parameters when debugging the code, such as the constraint tolerance in \emph{PJ} and the maximum radius of the chart validity region in \emph{AT} and \emph{TB}, which puts extra burdens on the developers.

From Table III, we know that the proposed framework outperforms \emph{PJ}-like centralized frameworks significantly as we plan the motion of the object and the robots hierarchically while \emph{PJ}-like frameworks plan the motion of the system simultaneously. In addition, $\emph{RRTConnect}$ (or $\emph{RRT}$) shows the best performance when working with our framework, hence it is chosen as the default solver in the following experiments.

\subsubsection{Simulation 2} \label{section_exp_sim2}

\begin{table*}[ht]
  \label{table_planner2}
  \caption{Performance of the Fully Decoupled Hierarchical Framework \cite{hekmatfar2014cooperative}}
  \begin{center}
  \scriptsize
  \resizebox{\linewidth}{!}{
    \begin{tabular}{c|cccccccccccc}
    \hline \multirow{2}{*}{Task} & \multicolumn{2}{c|}{\emph{RRT} \cite{lavalle1998rapidly}}   & \multicolumn{2}{c|}{\emph{RRTConnect} \cite{kuffner2000rrt}}  & \multicolumn{2}{c|}{\emph{BKPIECE} \cite{csucan2009kinodynamic}}    & \multicolumn{2}{c|}{\emph{EST} \cite{hsu1997path}}    & \multicolumn{2}{c|}{\emph{STRIDE} \cite{gipson2013resolution}}    & \multicolumn{2}{c}{\emph{PRM} \cite{kavraki1996probabilistic}}   \\ \cline{2-13} 
      & \begin{tabular}[c]{@{}c@{}}Success/\\ Total\end{tabular} & \multicolumn{1}{c|}{Time(s)} & \begin{tabular}[c]{@{}c@{}}Success/\\ Total\end{tabular} & \multicolumn{1}{c|}{Time(s)} & \begin{tabular}[c]{@{}c@{}}Success/\\ Total\end{tabular} & \multicolumn{1}{c|}{Time(s)} & \begin{tabular}[c]{@{}c@{}}Success/\\ Total\end{tabular} & \multicolumn{1}{c|}{Time(s)} & \begin{tabular}[c]{@{}c@{}}Success/\\ Total\end{tabular} & \multicolumn{1}{c|}{Time(s)} & \begin{tabular}[c]{@{}c@{}}Success/\\ Total\end{tabular} & Time(s)    \\ \hline
      Figure \ref{fig_sim_scene}(a)                 & 10/10                                                    & 0.06$\pm$0.01                & 10/10                                                    & 0.05$\pm$0.02                    & 10/10                                                    & 0.77$\pm$0.35                    & 10/10                                                    & 0.08$\pm$0.04                    & 10/10                                                    & 0.07$\pm$0.03                    & 10/10                                                    & 20.01$\pm$0.00 \\
      Figure \ref{fig_sim_scene}(b)                 & 10/10                                                    & 0.04$\pm$0.01                & 10/10                                                    & 0.05$\pm$0.01                    & 10/10                                                    & 0.53$\pm$0.20                    & 10/10                                                    & 0.07$\pm$0.03                    & 10/10                                                    & 0.09$\pm$0.03                    & 10/10                                                    & 20.01$\pm$0.00 \\
      Figure \ref{fig_sim_scene}(c)                 & 0/10                                                     & ---                          & 0/10                                                     & ---                          & 0/10                                                     & ---                          & 0/10                                                     & ---                          & 0/10                                                     & ---                          & 0/10                                                     & ---        \\
      Figure \ref{fig_sim_scene}(d)                 & 0/10                                                     & ---                          & 0/10                                                     & ---                          & 0/10                                                     & ---                          & 0/10                                                     & ---                          & 0/10                                                     & ---                          & 0/10                                                     & ---        \\
      Figure \ref{fig_sim_scene}(e)                 & 0/10                                                     & ---                          & 0/10                                                     & ---                          & 0/10                                                     & ---                          & 0/10                                                     & ---                          & 0/10                                                     & ---                          & 0/10                                                     & ---        \\
      Figure \ref{fig_sim_scene}(f)                 & 0/10                                                     & ---                          & 0/10                                                     & ---                          & 0/10                                                     & ---                          & 0/10                                                     & ---                          & 0/10                                                     & ---                          & 0/10                                                     & ---        \\ \hline
  \end{tabular}}
  \end{center}
\end{table*}

\begin{table*}[ht]
  \label{table_different_thres}
  \caption{Performance of Different \emph{thres} in the Semi-coupled Hierarchical Framework}
  \begin{center}
  \scriptsize
  \resizebox{\linewidth}{!}{
    \begin{tabular}{c|cccccccccccc}
    \hline \multirow{2}{*}{Task/\emph{thres}} & \multicolumn{2}{c|}{0.0}   & \multicolumn{2}{c|}{0.2}  & \multicolumn{2}{c|}{0.4}    & \multicolumn{2}{c|}{0.6}    & \multicolumn{2}{c|}{0.8}    & \multicolumn{2}{c}{1.0}   \\ \cline{2-13} 
      & \begin{tabular}[c]{@{}c@{}}Success/\\ Total\end{tabular} & \multicolumn{1}{c|}{Time(s)} & \begin{tabular}[c]{@{}c@{}}Success/\\ Total\end{tabular} & \multicolumn{1}{c|}{Time(s)} & \begin{tabular}[c]{@{}c@{}}Success/\\ Total\end{tabular} & \multicolumn{1}{c|}{Time(s)} & \begin{tabular}[c]{@{}c@{}}Success/\\ Total\end{tabular} & \multicolumn{1}{c|}{Time(s)} & \begin{tabular}[c]{@{}c@{}}Success/\\ Total\end{tabular} & \multicolumn{1}{c|}{Time(s)} & \begin{tabular}[c]{@{}c@{}}Success/\\ Total\end{tabular} & Time(s)    \\ \hline
      Figure \ref{fig_sim_scene}(a)                 & 10/10     & 0.82$\pm$0.24                & 10/10            & 1.01$\pm$0.44                    & 10/10             & 0.97$\pm$0.10                & 10/10          & 4.17$\pm$2.66                & 0/10              & ---                   & 0/10                    & ---                 \\
      Figure \ref{fig_sim_scene}(b)                 & 10/10     & 1.16$\pm$1.91                & 10/10            & 2.43$\pm$2.16                    & 10/10             & 3.34$\pm$1.56                & 10/10          & 6.16$\pm$2.30                & 4/10              & 21.85$\pm$14.07       & 0/10                    & ---                 \\
      Figure \ref{fig_sim_scene}(c)                 & 2/10      & 3.25$\pm$0.48                & 9/10             & 4.04$\pm$1.72                    & 10/10             & 6.99$\pm$5.71                & 10/10          & 12.22$\pm$3.09               & 2/10              & 32.08$\pm$11.28       & 0/10                    & ---                 \\
      Figure \ref{fig_sim_scene}(d)                 & 3/10      & 2.05$\pm$0.39                & 10/10            & 2.50$\pm$0.49                    & 10/10             & 7.98$\pm$7.06                & 10/10          & 16.68$\pm$6.97               & 3/10              & 33.83$\pm$13.26       & 0/10                    & ---                 \\
      Figure \ref{fig_sim_scene}(e)                 & 0/10      & ---                          & 5/10             & 16.54$\pm$4.87                   & 10/10             & 27.09$\pm$12.72              & 2/10           & 53.91$\pm$6.07               & 0/10              & ---                   & 0/10                    & ---                 \\
      Figure \ref{fig_sim_scene}(f)                 & 2/10      & 4.93$\pm$2.32                & 9/10             & 5.55$\pm$2.87                    & 10/10             & 5.73$\pm$1.91                & 1/10           & 52.32$\pm$0.00               & 0/10              & ---                   & 0/10                    & ---                 \\ \hline
  \end{tabular}}
  \end{center}
\end{table*}

\begin{figure}[ht]
  \centering
  \includegraphics[width=7cm]{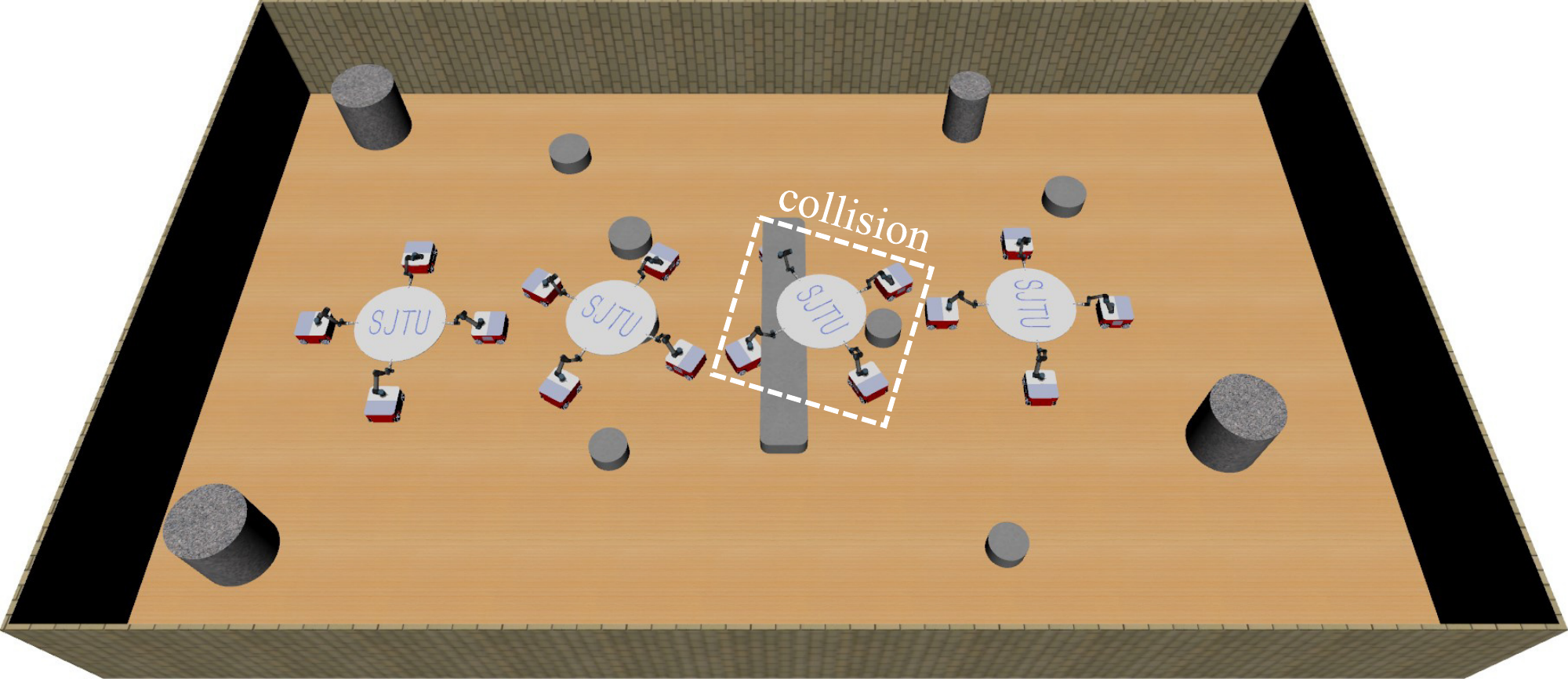}
  \caption{Robot path planned by the fully decoupled hierarchical framework \cite{hekmatfar2014cooperative}. Although the motion of the object is collision-free, \cite{hekmatfar2014cooperative} is unable to find a valid path that satisfies the collision, closed-chain and redundancy constraints simultaneously.}
  \label{fig_FDHF}
\end{figure}

In this simulation, we compare the proposed framework with the fully decoupled hierarchical framework. When the closed-chain constraint and the lower bound of the redundancy constraint metric are not guaranteed in Figure \ref{fig_overview}, the centralized layer and the decentralized layer become fully decoupled, resulting in a framework similar to \cite{hekmatfar2014cooperative}. The performance of the decoupled framework in the tasks in Figure \ref{fig_sim_scene}(a)-(f) is shown in Table IV.

When the scenes and the tasks are simple, the centralized layer and the decentralized layer are always compatible, so checking the closed-chain constraint and the lower bound of the redundancy constraint metric can be avoided. Therefore, the decoupled framework performs well in Figure \ref{fig_sim_scene}(a)-(b). However, obstacles will complicate the motion planning problem and make the two layers in conflict. Therefore, in Figure \ref{fig_sim_scene}(c)-(f), the performance of the decoupled framework decreases dramatically. A failure case is shown in Figure \ref{fig_FDHF}. As can be seen, although the motion of the object is collision-free, \cite{hekmatfar2014cooperative} cannot find a valid path for the robots that satisfies the collision, closed-chain and redundancy constraints simultaneously.

\subsubsection{Simulation 3} \label{section_exp_sim3}

\begin{figure}[ht]
  \centering
  \includegraphics[width=7cm]{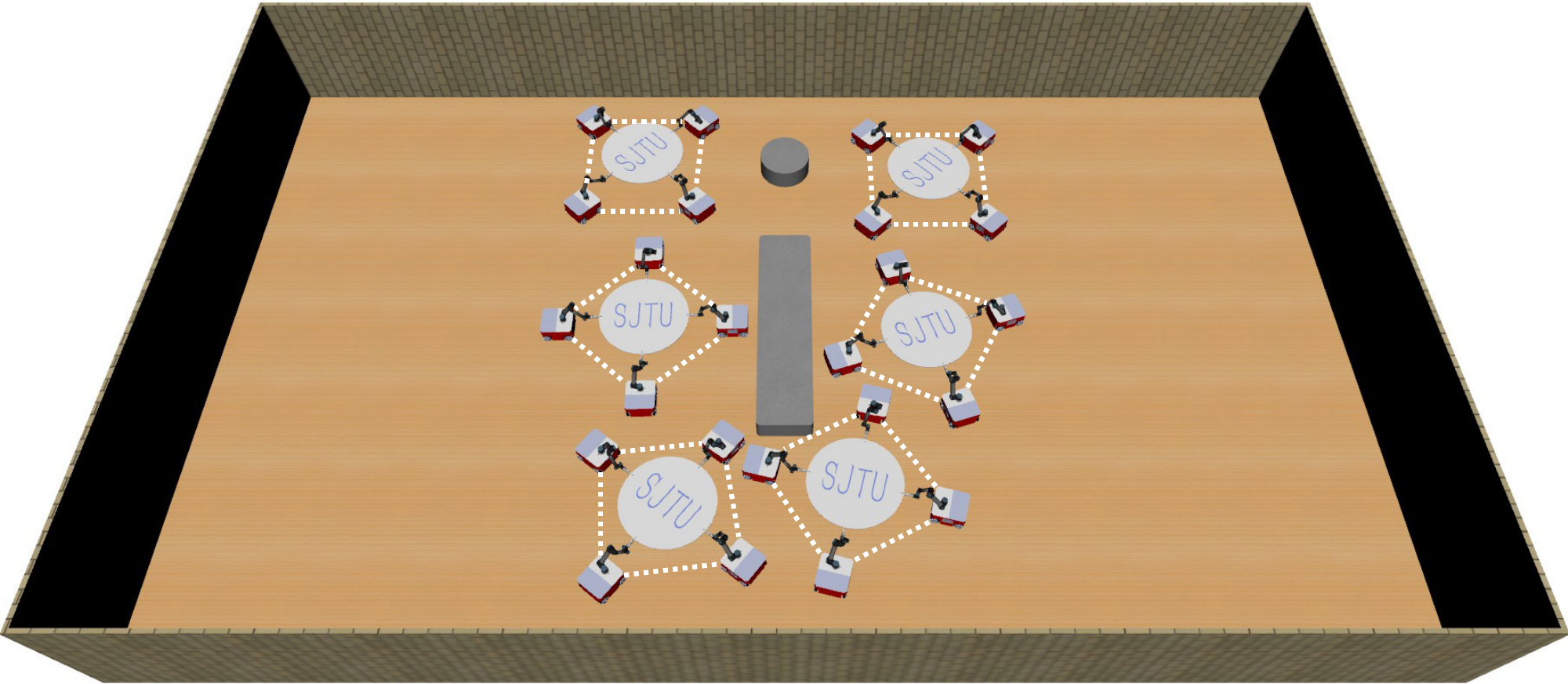}
  \caption{Robot path planned by algorithms based on the virtual structure \cite{tang2018obstacle, alonso2017multi, jiao2015transportation}. The dotted lines represent the virtual polygon that encircles the robots and the object. Compared with the path in Figure \ref{fig_sim_scene}(b), the proposed motion planner is able to find a shorter path that crosses obstacles, while \cite{tang2018obstacle, alonso2017multi, jiao2015transportation} only find a longer path that bypasses obstacles.}
  \label{fig_sceneB_SOR}
\end{figure}

In this simulation, we compare the proposed framework with algorithms based on the virtual structure. In \cite{tang2018obstacle, alonso2017multi, jiao2015transportation}, the object and the robots are encircled by polygons on the ground. As long as the virtual structure intersects with obstacles, the system is considered in collision. Therefore, the obstacle-avoidance strategy is conservative. Take the task in Figure \ref{fig_sim_scene}(b) as an example. Our motion planner is able to find a shorter path that crosses obstacles. However, the path in Figure \ref{fig_sim_scene}(b) becomes infeasible for \cite{tang2018obstacle, alonso2017multi, jiao2015transportation}, and they can only find a longer path that bypasses obstacles, which is shown in Figure \ref{fig_sceneB_SOR}. Crossing obstacles is an essential ability for the multiple robots system and will increase the success rate of the motion planner in cluttered environments.

\subsubsection{Simulation 4} \label{section_exp_sim4}

In this simulation, we test the effects of the coupling factor \emph{thres} on the \emph{SCHF}. The time cost and success rate in different tasks are shown in Table V and explained in the following.

When \emph{thres}=0, \emph{SCHF} performs well in Figure \ref{fig_sim_scene}(a)-(b). As the scenes become cluttered, such as Figure \ref{fig_sim_scene}(c)-(f), its performance is poor as the ``soft'' redundancy constraint is not considered in the centralized layer, which may cause conflict with the decentralized layer. Notice that closed-chain and collision of the \emph{MM} (condition \ding{174} in Eq. (\ref{eq_problem_definition2})) are always checked in this experiment, hence the time cost in Figure \ref{fig_sim_scene}(a)-(b) is slightly over the fully decoupled framework (Table IV).

When \emph{thres} is large, i.e., 0.8 and 1.0, the performance of the \emph{SCHF} is extremely poor and the reasons are twofolds. On the one hand, as \emph{thres} grows, so does the time cost as the centralized layer has to spend more time to find valid samples when planning the object's motion. As a result, the success rate will decrease within the limitted planning time. On the other hand, $f_{{\sss RC},i}(\boldsymbol{q}_i)$ is a relative normalized index w.r.t its global optimum in the entire configuration space. When obstacle and closed-chain exist, the robots may find locally optimal joint configurations but are hard to reach global optimums, hence the success rate will be 0 when \emph{thres}=1.0. Furthermore, \emph{thres} is also affected by the heterogeneity of the robots. For example, the system in Figure \ref{fig_sim_scene}(b) only includes the same \emph{MM}s, while Figure \ref{fig_sim_scene}(a) contains \emph{MM} with different mechanical structure.

In summary, \emph{thres} is affected by the definition of $f_{{\sss RC},i}(\boldsymbol{q}_i)$, the clutter of the environment/task and the heterogeneity of the robots. This simulation evaluates a possible expression of $f_{{\sss RC},i}(\boldsymbol{q}_i)$ and also give a guideline to choose a suitable \emph{thres} for other kind of $f_{{\sss RC},i}(\boldsymbol{q}_i)$ and tasks. A thumb rule is that in scenarios with no or few obstacles, if the requirement of planning time is critical, \emph{thres} can be set to a small value. In other scenarios, it should be set to a moderate value, e.g., 0.4, to balance the time cost and success rate in the \emph{SCHF}.

\subsection{Real-World Experiments} \label{section_exp_real}

In the real-world experiments, all robots are equipped with wireless transmission modules in the same local network for efficient communication. Leader-follower approach is adopted for the system, where the leader plans the global path of the object and each follower is equipped with a 6D force-torque sensor and a visual sensor on the end effector to detect the motion relative to the leader. Moreover, each mobile robot is equipped with a planar laser scan to detect the static and dynamic obstacles on the ground. In addtion, a coordinated motion controller proposed in \cite{zhang2019task} is used to track the motion from the planner. Before the transportation task, the control gain for each \emph{MM} should be adjusted independently to make sure that the mobile base and the end effector could track the desired motion timely and precisely.

\subsubsection{Experiment 1}
\begin{figure}[ht]
  \centering
  \includegraphics[width=8cm]{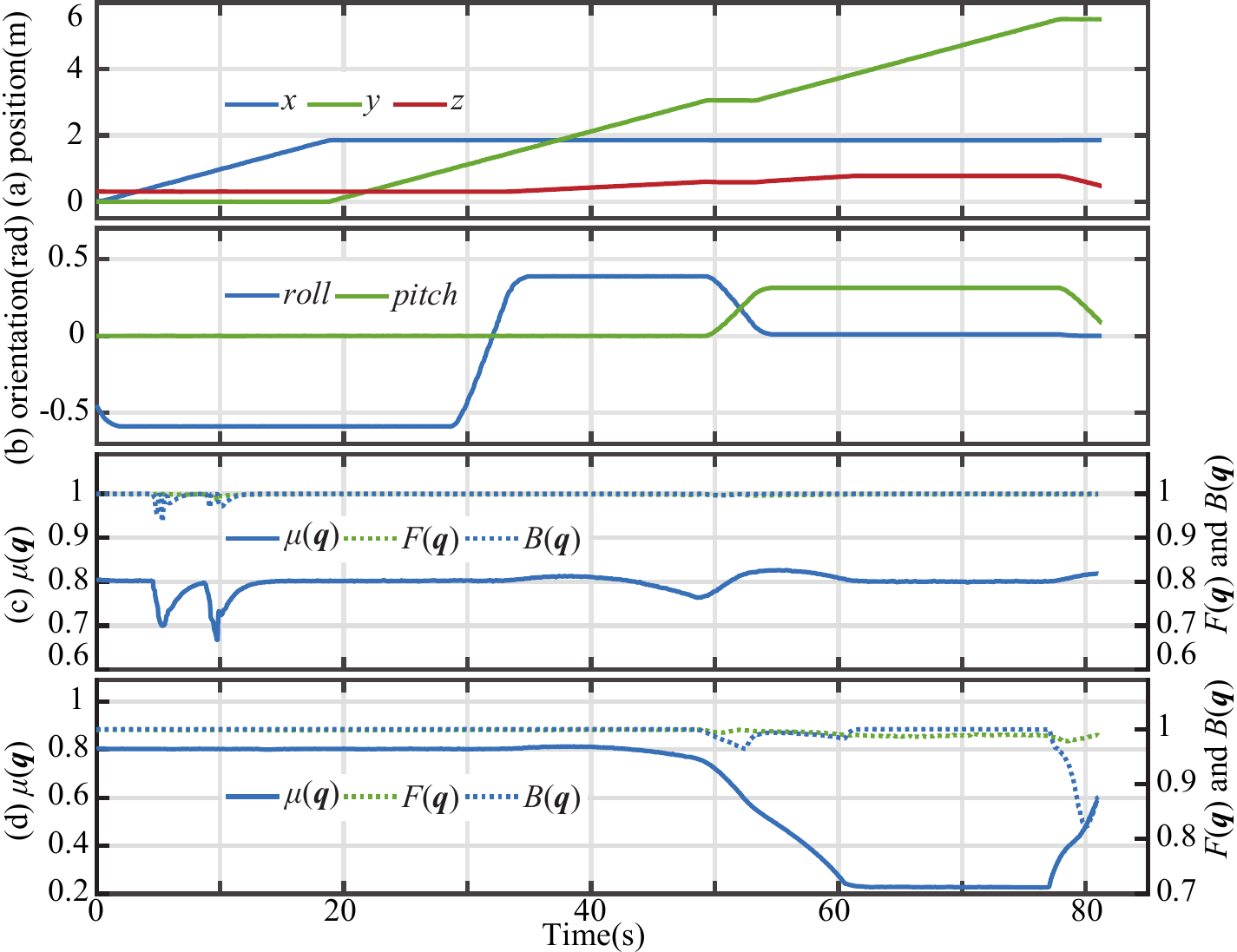}
  \caption{(a)-(b) Trajectories of the object in the task shown in Figure \ref{fig_real_two_robots}.  (c)-(d) $\mu(\boldsymbol{q})$, $F(\boldsymbol{q})$ and $B(\boldsymbol{q})$ of \emph{MM1} and \emph{MM2} during the task. }
  \label{fig_real_two_robots_data} 
\end{figure}

\begin{figure}[ht]
  \centering
  \includegraphics[width=7cm]{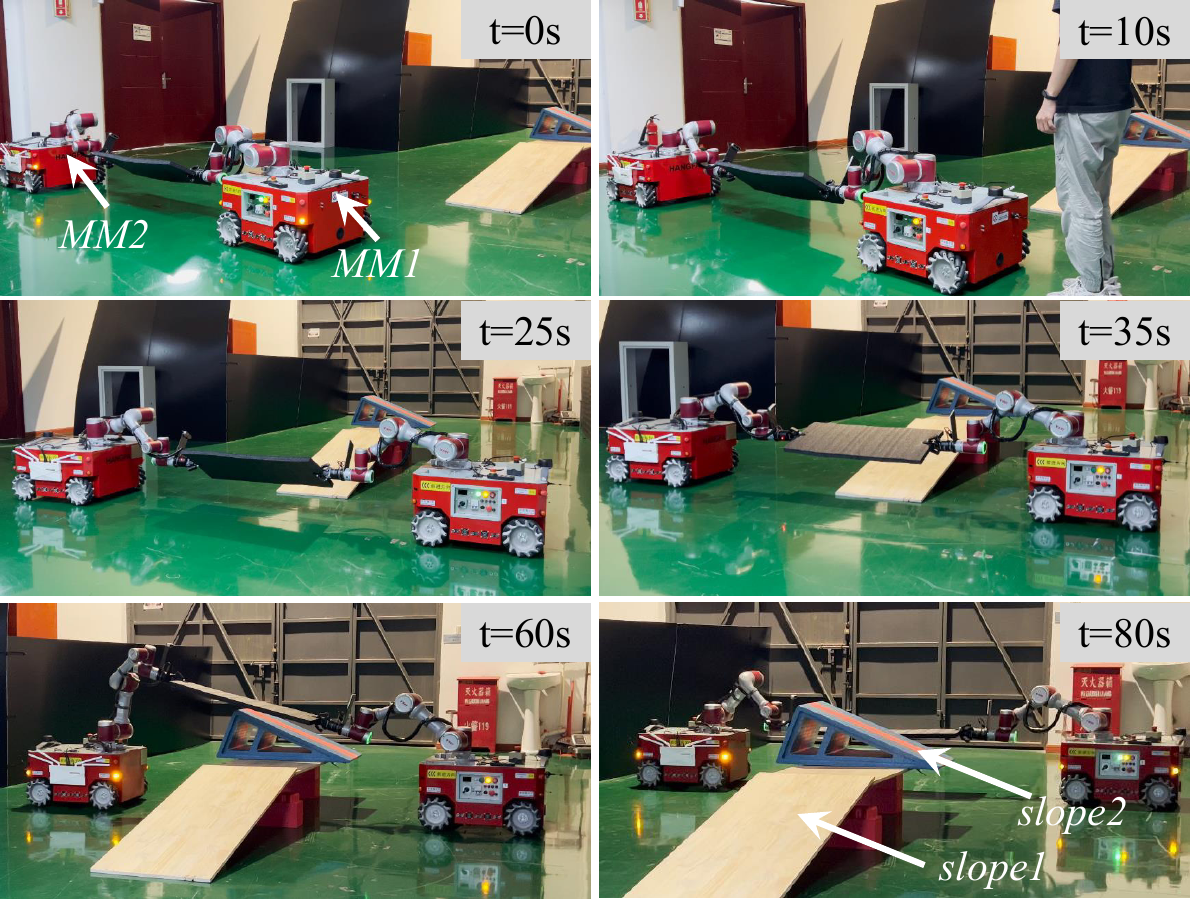}
  \caption{Scene of the real-world transportation task by two robots.}
  \label{fig_real_two_robots}
\end{figure}
In this experiment, we extend the proposed framework to the real-world cooperative transportation task. Obstacles include a pedestrian and two slopes with varying angles and heights. The system should transport the object to the destination while avoiding static and dynamic obstacles. The decentralized layer will optimize the joint configuration of the robots in real-time. Snapshots during the task are shown in Figure \ref{fig_real_two_robots}. The trajectories of the object, $\mu(\boldsymbol{q})$, $F(\boldsymbol{q})$ and $B(\boldsymbol{q})$ of \emph{MM1} and \emph{MM2} are shown in Figure \ref{fig_real_two_robots_data}. The motion process is explained as follows.
\begin{itemize}[leftmargin=*]
  \item At \emph{t=0s}, the system is far away from obstacles. Both \emph{MM1} and \emph{MM2} start with an optimal configuration.
  \item From \emph{t=5s} to \emph{t=10s}, \emph{MM1} meets the pedestrian twice. As a redundant robot, it is able to avoid the dynamic obstacle while not affecting the transportation task. As a punishment, $\mu(\boldsymbol{q})$ of \emph{MM1} is sacrificed (Figure \ref{fig_real_two_robots_data}(c)). However, when the obstacle moves away, $\mu(\boldsymbol{q})$ can be recovered to a high value.
  \item From \emph{t=28s} to \emph{t=35s}, the system adjusts \emph{z} and \emph{roll} of the object to suit the height and direction of \emph{slope1}. Thanks to the real-time redundancy resolution, $\mu(\boldsymbol{q})$, $F(\boldsymbol{q})$ and $B(\boldsymbol{q})$ of the two robots remain at high values during this period.
  \item From \emph{t=49s} to \emph{t=55s}, the height of the \emph{MM1}'s and \emph{MM2}'s end effector decreases and increases, respectively, which means that the system adjusts \emph{pitch} of the object to suit the height and direction of \emph{slope2}. During this period, \emph{MM2} gradually approaches the boundary of the workspace in \emph{z} direction, hence $\mu(\boldsymbol{q})$ decreases quickly (Figure \ref{fig_real_two_robots_data}(d)).
  \item From \emph{t=75s} to \emph{t=80s}, the system has crossed all obstacles and then lowers the object to finish the autonomous cooperative transportation task. Finally, the height of the \emph{MM2}'s end effector decreases, and $\mu(\boldsymbol{q})$ returns to a high value.
\end{itemize}
Therefore, multiple robots are able to cooperate with each other to realize the transportation task in cluttered environments, and each robot could plan the optimal joint configuration and track the desired trajectories precisely.

\subsubsection{Experiment 2}

\begin{figure}[ht]
  \centering
  \includegraphics[width=7cm]{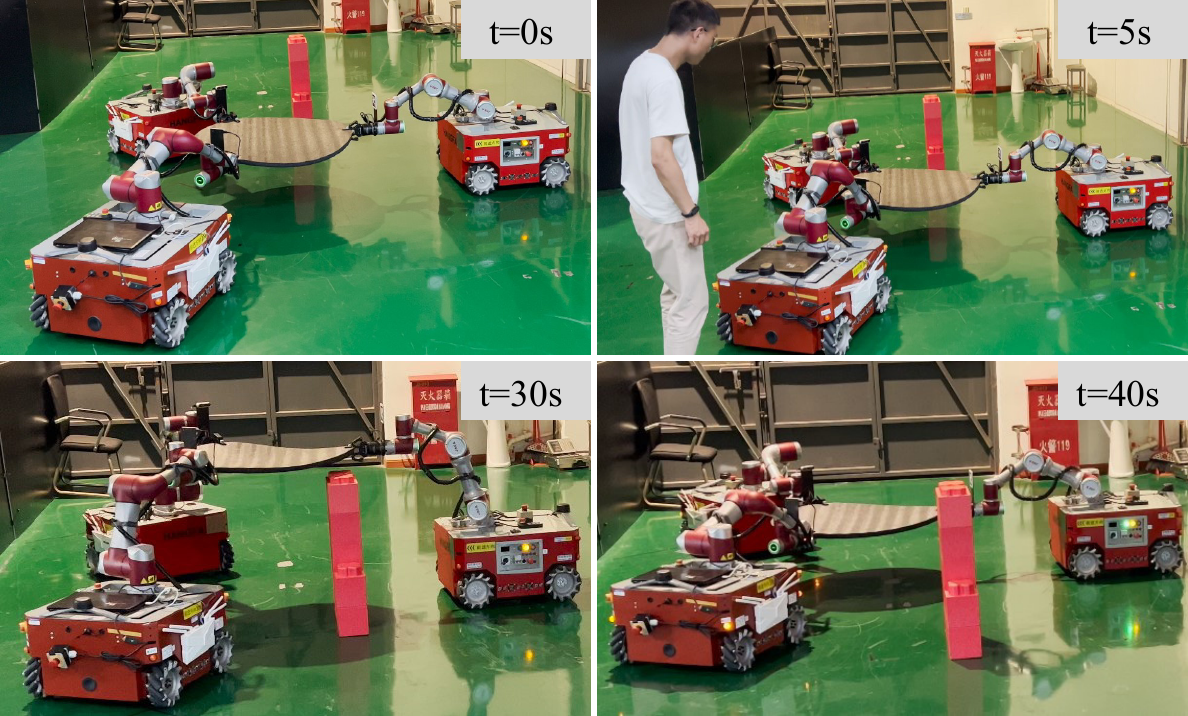}
  \caption{Scene of the real-world transportation task by three robots.}
  \label{fig_real_three_robots}
\end{figure}

In this experiment, we conduct the cooperative transportation task by three robots in real-world environment. Obstacles include a pedestrian and a cuboid with a certain height. Other settings are similar to the former experiment. Snapshots during the task are shown in Figure \ref{fig_real_three_robots}, and the pictorial motion is available in the attached video. From this and the former experiments, we can see that multiple robots are able to cooperate with each other to realize the transportation task in cluttered environments. Moreover, each robot could plan the optimal joint configuration and track the desired trajectories precisely.

\subsection{Discussions}

As can be seen from Figure \ref{fig_overview} and Eq. (\ref{eq_problem_definition2})-(\ref{eq_problem_definition3}), the proposed framework is semi-coupled, which means that we guarantee the closed-chain constraint $f_{\sss C^3}(\boldsymbol{c})$ and a user-specified lower bound of the redundancy constraint metric $f_{\sss RC}(\boldsymbol{c})$ in the centralized layer, and then optimize $f_{\sss RC}(\boldsymbol{c})$ on the basis of obeying $f_{\sss C^3}(\boldsymbol{c})$. Decoupling arises among different robots, which means that the decentralized layer is fully distributed, and the redundancy of each robot can be explored independently in real-time. 

Compared with the fully centralized frameworks, which use the numerical projection operator to generate valid samples and plan the motion of the object and the robots simultaneously, the hierarchical feature simplifies the motion planning problem of the high-dimensional system, which is verified in Section \ref{section_exp_sim1}. Compared with the decoupled frameworks, which plan the motion of the object and the robots separately, coupling the two layers can avoid conflict between them and enhance the performance of the motion planner, which is verified in Section \ref{section_exp_sim2}. Compared with the virtual structure-based algorithms, the proposed motion planner exhibits diversified and efficient obstacle-avoidance skills, such as bypassing and crossing, which is also verified in Section \ref{section_exp_sim3}.

The proposed framework can be applied to different numbers of heterogeneous mobile manipulators. To the best of our knowledge, such complex applications have not been seen before in the robotic community. However, there are several limitations in our work. Firstly, joint-limit constraint is neglected, and this is unreasonable for some complex tasks like flipping a sofa from up to down. Secondly, only kinematic constraints are considered. However, dynamic constraints, like inertia, slippage and deformation between the end effector and the object are essential in the real-world transportion tasks, which also call for more intelligent algorithm to plan and control the robots. Lastly, in the real-world experiment, each mobile robot has a localization error, which will cause relative motion among the end effectors of different robots, resulting in object deformation, and this will serve as an important direction of our future work.

\section{CONCLUSIONS} \label{section_conclusion}

In this paper, we propose a novel semi-coupled hierarchical framework to deal with the motion planning of multiple mobile manipulators and fulfill the cooperative transportation tasks in cluttered environment. The motion of the object is planned in the centralized layer offline. In the decentralized layer, the redundancy of each robot is explored independently in real-time. A user-specified lower bound of the redundancy constraint metric, closed-chain and obstacle-avoidance constraints are guaranteed in the centralized layer to ensure the object’s motion is executable in the decentralized layer. The proposed framework simplifies the complex motion planning problem while not violating the constraints. Its success rate and time cost outperform the benchmark motion planners significantly and can be applied to different numbers of heterogeneous mobile manipulators, which are verified in the simulated and real-world experiments.

\section*{APPENDIX} \label{Appendix}

\subsection*{A. Projection Operation for the Centralized Frameworks}

\IncMargin{1em}
\begin{algorithm}
\caption{\emph{Projection Through Iteration}}\label{alg_appen}
  \SetKwProg{Fn}{Function}{}{}
  \Fn{\texttt{\emph{projection}}($\boldsymbol{c}, \varepsilon$)}
  {
    \tcp{constraint error by Eq.(\ref{eq_ccc1})}
    $\Delta\boldsymbol{x}$ $\leftarrow$ $f_{\sss C^3}(\boldsymbol{c})$\;
    \While{$\|\Delta\boldsymbol{x}\| > \varepsilon$}
    {
        $\Delta\boldsymbol{c} \leftarrow \boldsymbol{J}_{\sss C^3}^{\dagger}(\boldsymbol{c})\Delta\boldsymbol{x}$\;
        $\boldsymbol{c} \leftarrow \boldsymbol{c} - \Delta\boldsymbol{c}$\;
        $\Delta\boldsymbol{x}$ $\leftarrow$ $f_{\sss C^3}(\boldsymbol{c})$\;
    }
  }
\end{algorithm}\DecMargin{1em}

Projection is the basis for the centralized frameworks \cite{yakey2001randomized}\cite{berenson2011task}\cite{jaillet2017path}\cite{kim2016tangent} in Section \ref{section_exp_sim1}. Alg. \ref{alg_appen} shows the iterative procedure that uses the pseudo-inverse Jacobian matrix of the closed-chain constraint function $f_{\sss C^3}(\boldsymbol{c})$.

Given a random sample $\boldsymbol{c}$, Alg. \ref{alg_appen} computes the constraint error according to Eq.(\ref{eq_ccc1}) and compares it with the maximum allowed error $\varepsilon$. When the error is larger than $\varepsilon$, the pseudo-inverse of the constrained Jacobian matrix $\boldsymbol{J}_{\sss C^3}^{\dagger}(\boldsymbol{c})$ will be calculated to project the random sample by the gradient descent operation.

\subsection*{B. Constrained Jacobian Matrix}

In this part, we derive the constrained Jacobian matrix $\boldsymbol{J}_{\sss C^3}(\boldsymbol{c}) \in \mathbb{R}^{6n \times \left(\sum_{i=1}^{n}{(n_{a,i} + n_{b,i})+6}\right)}$ of the system.
\begin{equation}
\label{eq_app_1}
    \boldsymbol{J}_{\sss C^3}(\boldsymbol{c}) =  \frac{\partial{f_{\sss C^3}(\boldsymbol{c})}}{\partial{\boldsymbol{c}}}
                                              =  \frac{\partial{\boldsymbol{E}} - \partial{\boldsymbol{G}}}{\partial{\boldsymbol{c}}}
\end{equation}

According to the definition of $\boldsymbol{E}$ and $\boldsymbol{G}$ in Section \ref{section_model_ccc}, $\boldsymbol{E}$ has no relationship with $\boldsymbol{t}_{obj}^{w}$, and $\boldsymbol{G}$ has no relationship with $\boldsymbol{q}_i$. Therefore, Eq. (\ref{eq_app_1}) can be simplified to Eq. (\ref{eq_app_2}).
\begin{equation}
\label{eq_app_2}
    \frac{\partial{\boldsymbol{E}} - \partial{\boldsymbol{G}}}{\partial{\boldsymbol{c}}} = \left( \frac{\partial{\boldsymbol{E}}}{\partial{\boldsymbol{q}_1}}, ...,
                                                                                                  \frac{\partial{\boldsymbol{E}}}{\partial{\boldsymbol{q}_i}}, ...,
                                                                                                  \frac{\partial{\boldsymbol{E}}}{\partial{\boldsymbol{q}_n}},
                                                                                                 -\frac{\partial{\boldsymbol{G}}}{\partial{\boldsymbol{t}_{obj}^w}} \right)
\end{equation}
where $\frac{\partial{\boldsymbol{E}}}{\partial{\boldsymbol{q}_i}}$ and $\frac{\partial{\boldsymbol{G}}}{\partial{\boldsymbol{t}_{obj}^w}}$ are defined in Eq. (\ref{eq_app_3}) and Eq. (\ref{eq_app_4}).
\begin{align}
\label{eq_app_3}
   & \frac{\partial{\boldsymbol{E}}}{\partial{\boldsymbol{q}_i}} = \left( \frac{\partial{f_k(\boldsymbol{q}_1)}}{\partial{\boldsymbol{q}_i}}, ...,
                                                                           \frac{\partial{f_k(\boldsymbol{q}_i)}}{\partial{\boldsymbol{q}_i}}, ...,
                                                                           \frac{\partial{f_k(\boldsymbol{q}_n)}}{\partial{\boldsymbol{q}_i}} \right)^T   \\
\label{eq_app_4}
   & \frac{\partial{\boldsymbol{G}}}{\partial{\boldsymbol{t}_{obj}^w}} = \left( \frac{\partial{\boldsymbol{t}_{g,1}^w}}{\partial{\boldsymbol{t}_{obj}^w}}, ...,
                                                                                \frac{\partial{\boldsymbol{t}_{g,i}^w}}{\partial{\boldsymbol{t}_{obj}^w}}, ...,
                                                                                \frac{\partial{\boldsymbol{t}_{g,n}^w}}{\partial{\boldsymbol{t}_{obj}^w}} \right)^T
\end{align}
where $\frac{\partial{f_k(\boldsymbol{q}_j)}}{\partial{\boldsymbol{q}_i}}$ is given by Eq. (\ref{eq_app_5}).
\begin{equation}
\label{eq_app_5}
    \frac{\partial{f_k(\boldsymbol{q}_j)}}{\partial{\boldsymbol{q}_i}} = \begin{cases}
                                                                	      \boldsymbol{J}_i(\boldsymbol{q}_{i}), & \emph{if} \ i = j \\
                                                                	      \boldsymbol{O}_{6 \times (n_{a,i}+n_{b,i})}, & \emph{if} \ i\neq j	
                                                                		   \end{cases}
\end{equation}
where $\boldsymbol{J}_i(\boldsymbol{q}_{i})$ is the analytical Jacobian matrix of the $i$th \emph{MM}.

Suppose the constant homogeneous transformation of frame $O_{g,i}X_{g,i}Y_{g,i}Z_{g,i}$ relative to frame $O_{obj}X_{obj}Y_{obj}Z_{obj}$ is
$$    \boldsymbol{X}_{g,i}^{obj} = \left(
                                   \begin{array}{cc}
                                     \boldsymbol{R}_{g,i}^{obj} & \boldsymbol{p}_{g,i}^{obj} \\
                                     \boldsymbol{O}_{1\times3} & 1 \\
                                   \end{array}
                                 \right).$$
Suppose the velocity of a frame is $\boldsymbol{\xi} = (\boldsymbol{v}^T, \boldsymbol{\omega}^T)^T$, in which $\boldsymbol{v}$ and $\boldsymbol{\omega}$ represent the linear and angular velocity of the frame. According to \cite{caccavale2016cooperative}, Eq. (\ref{eq_app_6}) holds between $\boldsymbol{\xi}_{g,i}^w$ and $\boldsymbol{\xi}_{obj}^w$.
\begin{equation}
\label{eq_app_6}
    \boldsymbol{\xi}_{g,i}^w = \left(
                                   \begin{array}{cc}
                                     \boldsymbol{I}_{3\times3} & -S(\boldsymbol{p}_{g,i}^{w}) \\
                                     \boldsymbol{O}_{3\times3} & \boldsymbol{I}_{3\times3} \\
                                   \end{array}
                                 \right)
                                 \boldsymbol{\xi}_{obj}^w
\end{equation}
where $S(.)$ is the skew-symmetric matrix operator. In addition, we use a minimum description to represent the orientation in $\boldsymbol{t}$, hence Eq. (\ref{eq_app_7}) holds between $\dot{\boldsymbol{t}}$ and $\boldsymbol{\xi}$.
\begin{equation}
\label{eq_app_7}
    \boldsymbol{\xi} = \left(
                                   \begin{array}{cc}
                                     \boldsymbol{I}_{3\times3} & \boldsymbol{O}_{3\times3} \\
                                     \boldsymbol{O}_{3\times3} & B(\boldsymbol{\alpha}) \\
                                   \end{array}
                                 \right)
                                 \dot{\boldsymbol{t}}
\end{equation}
When \emph{roll-pitch-yaw} angle is used to represent the orientation, we have $\boldsymbol{\alpha} = (\phi, \psi, \theta)^T$ and $B(\boldsymbol{\alpha}) = \left(
                                                                                                                          \begin{array}{ccc}
                                                                                                                            c_{\psi}c_{\theta} & -s_{\theta} & 0 \\
                                                                                                                            c_{\psi}s_{\theta} & c_{\theta} & 0 \\
                                                                                                                            -s_{\psi} & 0 & 1 \\
                                                                                                                          \end{array}
                                                                                                                        \right)$,
where $s$ and $c$ represent sine and cosine operator, respectively.

Combining Eq. (\ref{eq_app_6}) and Eq. (\ref{eq_app_7}), $\frac{\partial{\boldsymbol{t}_{g,i}^w}}{\partial{\boldsymbol{t}_{obj}^w}}$ can be derived in Eq. (\ref{eq_app_8}).
\begin{equation}
\label{eq_app_8}
    \frac{\partial{\boldsymbol{t}_{g,i}^w}}{\partial{\boldsymbol{t}_{obj}^w}} = \boldsymbol{W}_{i} =
                                 \left(
                                   \begin{array}{cc}
                                     \boldsymbol{I}_{3\times3} & -S(\boldsymbol{p}_{g,i}^{w})B(\boldsymbol{\alpha}_{obj}^w) \\
                                     \boldsymbol{O}_{3\times3} & -B^{-1}(\boldsymbol{\alpha}_{g,i}^w)B(\boldsymbol{\alpha}_{obj}^w) \\
                                   \end{array}
                                 \right)
\end{equation}
Combining Eq. (\ref{eq_app_1})-(\ref{eq_app_5}) and Eq. (\ref{eq_app_8}), $\boldsymbol{J}_{\sss C^3}(\boldsymbol{c})$  is finally given in Eq. (\ref{eq_app_9}).
\begin{equation}
\label{eq_app_9}
   \boldsymbol{J}_{\sss C^3}(\boldsymbol{c})  = \left(
                                                  \begin{array}{cccc}
                                                    \boldsymbol{J}_1(\boldsymbol{q}_{1})    & ...       & \boldsymbol{O}                        & -\boldsymbol{W}_{0} \\
                                                    \vdots                                  & \ddots    & \vdots                                & \vdots \\
                                                    \boldsymbol{O}                          & ...       & \boldsymbol{J}_n(\boldsymbol{q}_{n})  & -\boldsymbol{W}_{n} \\
                                                  \end{array}
                                                \right)
\end{equation}

\subsection*{C. Time and Space Complexity Analysis}

According to Alg. \ref{alg_appen}, the most time consuming operation in the iteration process is calculating the pseudo-inverse of the Jacobian matrix. According to the SVD algorithm \cite{li2019tutorial}, for a matrix with dimension $p \times q$, the time complexity and space complexity when calculating the pseudo-inverse can be approximated to $\mathcal{O}_t(pq^2)$ and $\mathcal{O}_s(pq)$.

In the fully centralized framework, $\boldsymbol{J}_{\sss C^3}(\boldsymbol{c})$ should be used in Alg. \ref{alg_appen}, and the dimension is $p=6n$ and $q=\sum_{i=1}^{n}{(n_{a,i} + n_{b,i})+6}$. Since $p$ and $q$ are linearly related to $n$, $\mathcal{O}_t(pq^2)$ and $\mathcal{O}_s(pq)$ can be simplified to $\mathcal{O}_t(n^3)$ and $\mathcal{O}_s(n^2)$, respectively.

In the semi-coupled and fully decoupled framework, $\boldsymbol{J}_i(\boldsymbol{q}_{i}) (i=1,...,n)$ will be used in the iteration process. The dimension of each matrix is $p=6$ and $q=n_{a,i} + n_{b,i}$, respectively. Due to $p$ and $q$ are constant, $\mathcal{O}_t(pq^2)$ and $\mathcal{O}_s(pq)$ can be simplified to $\mathcal{O}_t(n)$ and $\mathcal{O}_s(n)$, respectively.

\begin{con}
  \ctitle{Author Contributions}
  
  Zhenhua Xiong and Xiangyang Zhu led the project and proposed the core idea of this paper. Heng Zhang wrote the article and conducted the experiments. Xinjun Sheng, Haoyi Song and Wenhang Liu helped to conducted the expetiment, performed data analyses and proposed insightful discussions to this work. 
  
  \ctitle{Financial Support}
  This work was supported in part by the National Natural Science Foundation of China (U1813224, 51675325), and Ministry of Education China Mobile Research Fund Project (MCM20180703).
  
  \ctitle{Conflicts of Interest}
  There is no conflicts of interest exist.
  
  \ctitle{Ethical Approval}
  Not applicable.
  \end{con}

\end{document}